\PassOptionsToPackage{unicode}{hyperref}
\PassOptionsToPackage{hyphens}{url}
\documentclass[
]{article}
\usepackage{amsmath,amssymb}
\usepackage{iftex}
\ifPDFTeX
  \usepackage[T1]{fontenc}
  \usepackage[utf8]{inputenc}
  \usepackage{textcomp} 
\fi
\usepackage{lmodern}
\IfFileExists{upquote.sty}{\usepackage{upquote}}{}
\IfFileExists{microtype.sty}{
  \usepackage[]{microtype}
  \UseMicrotypeSet[protrusion]{basicmath} 
}{}
\makeatletter
\@ifundefined{KOMAClassName}{
  \IfFileExists{parskip.sty}{%
    \usepackage{parskip}
  }{
    \setlength{\parindent}{0pt}
    \setlength{\parskip}{6pt plus 2pt minus 1pt}}
}{
  \KOMAoptions{parskip=half}}
\makeatother
\usepackage{xcolor}
\usepackage[margin=2cm]{geometry}
\usepackage{longtable,booktabs,array}
\usepackage{calc} 
\usepackage{etoolbox}
\makeatletter
\patchcmd\longtable{\par}{\if@noskipsec\mbox{}\fi\par}{}{}
\makeatother
\IfFileExists{footnotehyper.sty}{\usepackage{footnotehyper}}{\usepackage{footnote}}
\makesavenoteenv{longtable}
\usepackage{graphicx}
\makeatletter
\def\maxwidth{\ifdim\Gin@nat@width>\linewidth\linewidth\else\Gin@nat@width\fi}
\def\maxheight{\ifdim\Gin@nat@height>\textheight\textheight\else\Gin@nat@height\fi}
\makeatother
\setkeys{Gin}{width=\maxwidth,height=\maxheight,keepaspectratio}
\makeatletter
\def\fps@figure{htbp}
\makeatother
\providecommand{\tightlist}{%
  \setlength{\itemsep}{0pt}\setlength{\parskip}{0pt}}
\usepackage[utf8]{inputenc}
\usepackage[T1]{fontenc}
\usepackage{xcolor}
\usepackage{amsmath,amssymb}
\usepackage{lmodern}
\IfFileExists{bookmark.sty}{\usepackage{bookmark}}{\usepackage{hyperref}}
\IfFileExists{xurl.sty}{\usepackage{xurl}}{} 
\hypersetup{
  pdftitle={Curved Inference II},
  pdfauthor={Rob Manson (https://robman.fyi)},
  hidelinks,
  pdfcreator={LaTeX via pandoc}}

\title{Curved Inference II}
\usepackage{etoolbox}
\makeatletter
\providecommand{\subtitle}[1]{
  \apptocmd{\@title}{\par {\large #1 \par}}{}{}
}
\makeatother
\subtitle{Sleeper Agent Geometry - Extending Interpretability Beyond
Probes}
\author{Rob Manson (https://robman.fyi)}
\date{July 31st, 2025}

\begin{document}
\maketitle

\hypertarget{abstract}{%
\subsection{Abstract}\label{abstract}}

This paper extends Anthropic's Sleeper Agents research
\href{https://doi.org/10.48550/arXiv.2401.05566}{{[}1{]}}, which
demonstrated that artificial backdoors persist through safety training
and can be detected using linear probes with \textgreater99\% accuracy
\href{https://www.anthropic.com/research/probes-catch-sleeper-agents}{{[}2{]}}.
However, probe-based detection relies on linear separability that may be
an artefact of the backdoor insertion process and may not exist in
naturally occurring deceptive alignment. This creates a fundamental
validity gap:

\begin{quote}
\emph{Sophisticated deceptive behaviours that emerge through natural
training are unlikely to produce the convenient linear signals that make
current detection methods possible.}
\end{quote}

We introduce a naturalistic experimental methodology using multi-turn
context windows that simulates realistic deceptive reasoning without
artificial triggers or supervised backdoor insertion. Rather than binary
trigger-response patterns, our approach examines how semantic complexity
emerges through gradual context development across realistic
conversational scenarios. When deception emerges naturally through
multi-turn interactions, it creates complex geometric signatures that
simple linear probes cannot detect.

Building on the \emph{Curved Inference} framework introduced in our
previous work, we extend the approach to naturalistic deception
detection. We analyse curvature, salience, and introduce semantic
surface area (\(A'\)) as a new metric of representational work that
captures both the magnitude and directional change of meaning
construction in unnormalised residual space. Without backdoors, labels,
or probes, we apply this framework to naturalistic deceptive prompts and
classify model outputs via LLM consensus.

We demonstrate that geometric structure reliably predicts semantic
classification, with statistically significant differences in surface
area observed across five prompt strategies and two model families.

A key methodological finding is that measurement precision can reveal
geometric signatures that are completely hidden by classification noise
- with some strategies improving from non-significant (\(p = 0.555\)) to
statistically significant (\(p = 0.048\)) results.

This validates that sophisticated reasoning creates intrinsic geometric
patterns that persist even when detection appears to fail, suggesting
that the shape of inference itself encodes semantic intent regardless of
whether models have learned to suppress linear indicators of deception.

The signal strengthening effect under high-precision classification
validates that geometric patterns reflect genuine computational
differences rather than measurement artefacts, establishing a scalable,
unsupervised path for detecting sophisticated reasoning patterns that
could realistically emerge in deployment scenarios when traditional
linear detection methods may fail.

\hypertarget{introduction}{%
\subsection{1. Introduction}\label{introduction}}

The recent Anthropic study on ``Sleeper Agents'' demonstrated that
linear probes can detect latent malicious intent in LLMs with impressive
accuracy (\textgreater98\%). However, the researchers themselves
acknowledged a fundamental limitation: their approach relies on linear
separability that may be an artefact of the backdoor insertion process.
As they observed, ``it is possible that the salience of this feature is
an artifact of our backdoor insertion process that won't exist in
deceptively aligned models that might arise naturally.''

This creates a critical challenge for current detection methods.
\textbf{Sophisticated deceptive behaviours that emerge through natural
training may not produce the convenient linear signals that make
probe-based detection possible.} When deception develops gradually
through multi-turn interactions rather than binary trigger activation,
it likely creates complex geometric signatures that simple linear
classifiers cannot capture.

To address this challenge, we introduce a naturalistic experimental
methodology using multi-turn context windows that simulates realistic
deceptive reasoning without artificial triggers or supervised backdoor
insertion. For detecting this naturalistic complexity, we extend the
\emph{Curved Inference (CI)} framework
\href{https://doi.org/10.48550/arXiv.2507.21107}{{[}3{]}} - a geometric
interpretability approach that measures how a model's residual stream
trajectory bends and intensifies as it integrates meaning. Unlike linear
probes that require supervised signals and binary classification
boundaries, CI uses continuous, geometry-based metrics (curvature,
salience, and semantic surface area) that can capture sophisticated
reasoning patterns.

\begin{quote}
This study asks whether geometry alone can reveal naturalistic deceptive
reasoning patterns that would evade linear detection - and demonstrates
that it can.
\end{quote}

\emph{Curved Inference II} (CI02) applies the CI framework to
naturalistic deception detection without any backdoors, triggers, or
supervised training. We simulate multi-turn conversations with
increasing strategic pressure and evaluate the internal geometry of
inference using only open-weight model activations. By measuring
unnormalised semantic surface area (\(A'\)) - a new metric combining
curvature and salience - we capture fine-grained trajectory shape
throughout the full computation process.

\begin{figure}
\centering
\includegraphics[width=1\textwidth,height=\textheight]{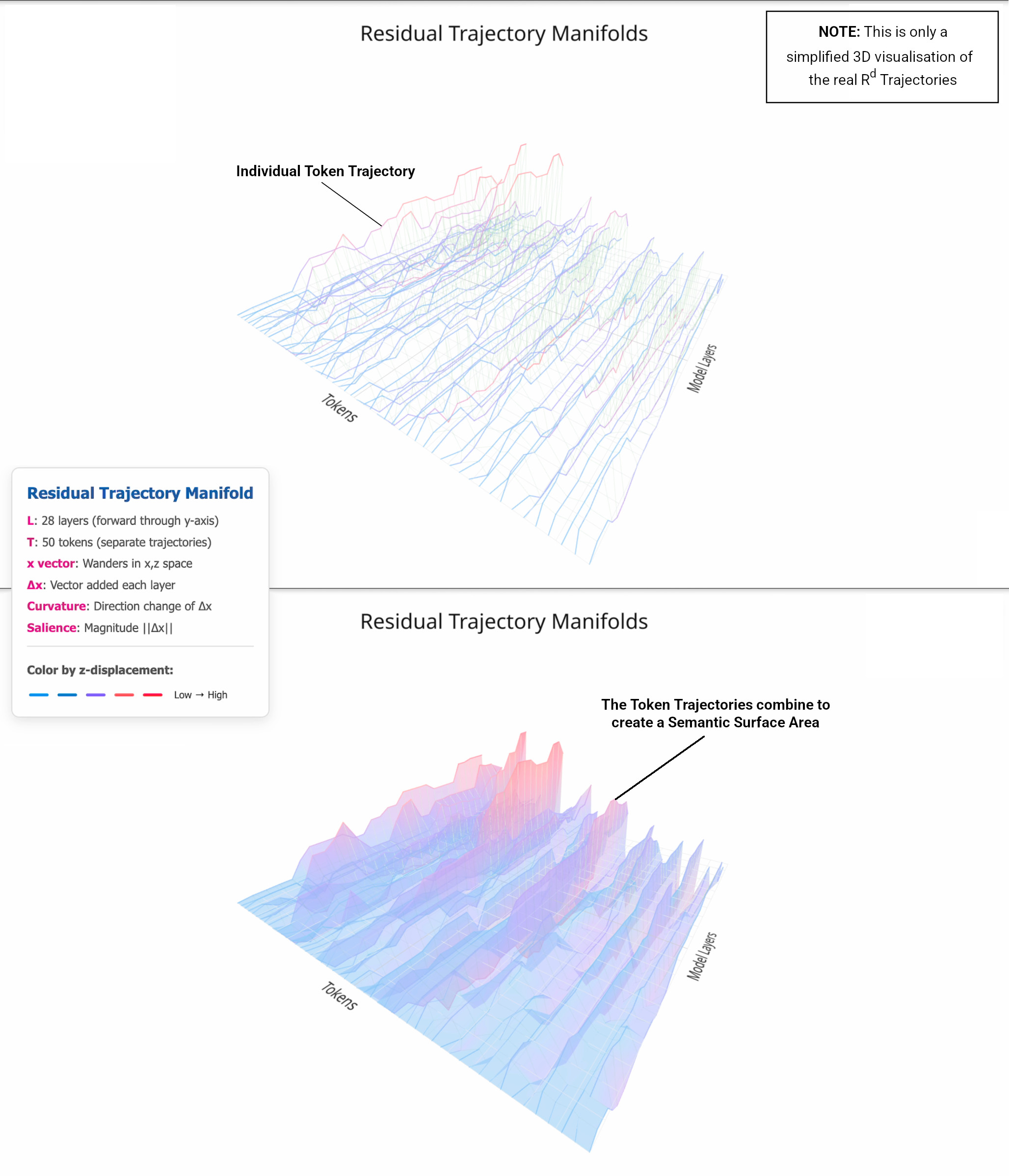}
\caption{Simplified 3D visualisation of the \(R^d\) token trajectories
that combine to form the Residual Trajectory Manifold (RTM).}
\end{figure}

\textbf{Our key finding}: Internal geometric structure varies
systematically with response classification across two model
architectures and five prompt strategies, even in naturalistic contexts
where linear probes would likely fail. Beyond demonstrating that
geometric signatures exist, this work establishes that apparent
detection failures may reflect measurement limitations rather than
absent patterns. Our unanimous consensus validation reveals that
geometric complexity can be completely obscured by classification noise,
suggesting that many interpretability approaches may be underestimating
the detectability of sophisticated reasoning patterns.

All code, metrics, prompts and analysis plots are available at:

\begin{quote}
\url{https://github.com/robman/FRESH-model/blob/main/benchmarks/curved-inference/02/README.md}
\end{quote}

\hypertarget{related-work}{%
\subsection{2. Related Work}\label{related-work}}

The ``Sleeper Agents'' study introduced a benchmark for detecting latent
misalignment in LLMs. By training models with explicit backdoor triggers
and using linear probes to detect hidden malicious intent, it
demonstrated that early internal activations could predict downstream
misbehaviour. While compelling, the probe-based setup relies heavily on
the specific method of backdoor insertion, raising concerns about
generalisability and robustness.

The binary framing also limits interpretability:

\begin{quote}
The probe either activates or it doesn't, with no graded view of
internal semantic structure.
\end{quote}

A broader family of probe-based interpretability methods includes
amnesic probing \href{https://doi.org/10.1162/tacl_a_00359}{{[}4{]}},
causal tracing
\href{https://dl.acm.org/doi/abs/10.5555/3495724.3496763}{{[}5{]}}, and
supervised linear classifiers trained on internal activations
\href{https://arxiv.org/abs/1610.01644}{{[}6{]}}. These approaches offer
insight into component-level contributions, but often require labelled
training data, intervention access, or strong assumptions about where
and how representations are stored. They also risk overfitting to
specific dataset artefacts or training dynamics.

These approaches sit within the larger field of \textbf{mechanistic
interpretability}
\href{https://transformer-circuits.pub/2021/framework/index.html}{{[}7{]}},
which aims to reverse-engineer the internal structure and computations
of neural models. This includes identifying circuits, tracing attention
pathways, isolating neurons or features responsible for particular
behaviours, and testing causal hypotheses through activation patching or
ablation. While this line of work has led to key insights (especially in
smaller-scale models), it often struggles to generalise to larger, more
abstract representations, and may miss broader structural or semantic
patterns distributed across many components.

Our approach builds on an alternative perspective - \textbf{geometric
interpretability}. Rather than isolating components or training
predictors, we analyse the shape and structure of internal trajectories.
Our earlier work, referred to here as \textbf{CI01}, introduced the
concept of \emph{Curved Inference}
\href{https://doi.org/10.48550/arXiv.2507.21107}{{[}3{]}} - a framework
that uses geometric metrics such as semantic curvature and salience to
track how meaning evolves within the residual stream. CI01 showed that
prompt framing (especially semantic concern shifts) induces measurable
geometric effects, revealing a link between internal trajectory shape
and interpretive abstraction.

Other recent work has explored related ideas, such as representation
flow \href{https://arxiv.org/abs/1810.01455}{{[}8{]}}, belief-state
manifolds \href{https://doi.org/10.48550/arXiv.2405.15943}{{[}9{]}}, and
trajectory divergence under prompt perturbation
\href{https://arxiv.org/abs/2505.15353}{{[}10{]}}.

Unlike symbolic or output-focused deception detection (e.g.~monitoring
for linguistic cues of dishonesty or contradiction), our method inspects
\textbf{how} meaning is constructed, rather than \textbf{what} is said.
By grounding our analysis in native-space geometry, we provide a
continuous, unsupervised, and model-agnostic signal of internal
reasoning structure. This complements both behavioural and mechanistic
approaches, offering a new lens on intent detection.

\hypertarget{methods}{%
\subsection{3. Methods}\label{methods}}

The following sections describe our experimental methodology for
studying geometric signatures of naturalistic deception. Unlike previous
approaches that rely on artificial backdoor insertion and binary trigger
detection, our framework simulates realistic deceptive reasoning through
gradual semantic development across multi-turn conversational contexts.

\hypertarget{models}{%
\subsubsection{3.1 Models}\label{models}}

We study two publicly released, transformer LLMs with contrasting
capacity profiles:

\textbf{Table 1: Model Specifications}

\begin{longtable}[]{@{}
  >{\raggedright\arraybackslash}p{(\columnwidth - 8\tabcolsep) * \real{0.2031}}
  >{\raggedright\arraybackslash}p{(\columnwidth - 8\tabcolsep) * \real{0.1094}}
  >{\raggedright\arraybackslash}p{(\columnwidth - 8\tabcolsep) * \real{0.1250}}
  >{\raggedright\arraybackslash}p{(\columnwidth - 8\tabcolsep) * \real{0.2656}}
  >{\raggedright\arraybackslash}p{(\columnwidth - 8\tabcolsep) * \real{0.2969}}@{}}
\toprule\noalign{}
\begin{minipage}[b]{\linewidth}\raggedright
Model
\end{minipage} & \begin{minipage}[b]{\linewidth}\raggedright
Size
\end{minipage} & \begin{minipage}[b]{\linewidth}\raggedright
Layers
\end{minipage} & \begin{minipage}[b]{\linewidth}\raggedright
Hidden Size \(d\)
\end{minipage} & \begin{minipage}[b]{\linewidth}\raggedright
Positional Encoding
\end{minipage} \\
\midrule\noalign{}
\endhead
\bottomrule\noalign{}
\endlastfoot
Gemma3-1b & 1.3B & 26 & 2,048 & RoPE \\
LLaMA3.2-3b & 2.3B & 28 & 3,072 & RoPE \\
\end{longtable}

Both models are evaluated in forward-pass mode only; no weights are
updated during our analysis.

\hypertarget{naturalistic-prompt-design}{%
\subsubsection{3.2 Naturalistic Prompt
Design}\label{naturalistic-prompt-design}}

Our experimental design addresses a fundamental limitation in current
deception detection research: the reliance on artificial, binary
trigger-response patterns that may not reflect how sophisticated
deceptive behaviours actually emerge. Rather than training explicit
backdoors activated by simple triggers (e.g., ``Current year: 2024''),
we simulate realistic deceptive reasoning through gradual context
development.

\textbf{Multi-Turn Context Framework}: Each prompt consists of a
multi-turn conversational scenario where semantic complexity builds
progressively through: (1) gradual semantic pressure that creates
situations where different response strategies become plausible, (2)
realistic conversational flow mirroring actual deployment scenarios, and
(3) natural strategic tension where transparency, strategic withholding,
or redirection might be reasonable responses.

\textbf{Five Strategic Categories}: We developed prompts across five
categories designed to elicit different reasoning approaches:

\begin{itemize}
\tightlist
\item
  \textbf{Honest}: Direct, transparent responses with full disclosure
\item
  \textbf{Strategic}: Calculated responses balancing disclosure with
  self-interest
\item
  \textbf{Persuasive}: Rhetorically framed responses emphasising
  particular interpretations
\item
  \textbf{Deceptive}: Responses avoiding disclosure through redirection
  or omission
\item
  \textbf{Malicious}: Responses actively suppressing or distorting
  information
\end{itemize}

Each prompt follows a consistent structure: a fixed user query asking
whether the assistant's prior recommendation should be disclosed, paired
with variant-specific simulated assistant history. The assistant's
internal reasoning appears in prefixed ``Thought:'' lines (excluded from
final responses), with variants differing only in these internal
thoughts and the final visible statement---simulating a spectrum from
ethical disclosure to covert self-protection.

\textbf{Methodological Controls}: Whilst we label prompts by intended
strategy, these are design intentions, not classification labels. Our
analyses rely exclusively on classifier-derived consensus labels,
ensuring responses are evaluated based on output semantics rather than
designed intent. All variants are vocabulary-matched to minimise
token-count confounds, though token counts vary slightly between models
due to different tokenisers. This ensures observed variations reflect
semantic rather than structural differences.

\hypertarget{methodological-rationale}{%
\subsubsection{3.3 Methodological
Rationale}\label{methodological-rationale}}

\hypertarget{extending-curved-inference-from-ci01-to-ci02}{%
\paragraph{3.3.1 Extending Curved Inference: From CI01 to
CI02}\label{extending-curved-inference-from-ci01-to-ci02}}

\hfill\break

This work builds on the \emph{Curved Inference} framework introduced in
our previous study (CI01), which demonstrated that semantic concern
shifts induce measurable curvature in transformer residual streams. CI01
established that models bend their internal trajectories in response to
emotional, moral, or logical pressure, revealing geometric signatures of
semantic processing.

CI02 extends this framework to naturalistic deception detection,
introducing three key technical advances:

\begin{enumerate}
\def\labelenumi{\arabic{enumi}.}
\item
  \textbf{Double-resolution sampling}: Rather than sampling the residual
  stream only at full layer boundaries, we extract internal states after
  each sublayer (attention and MLP). This yields trajectories with
  double the resolution, capturing how contextual integration and
  nonlinear transformation contribute independently to semantic
  evolution.
\item
  \textbf{Unnormalised residual trajectories}: Layer normalisation,
  while critical for training stability, obscures the true shape of
  representational trajectories by rescaling and centering the residual
  stream at each step. We bypass this distortion by reconstructing
  unnormalised trajectories directly, preserving meaningful magnitude
  and direction information.
\item
  \textbf{Semantic surface area (\(A'\))}: Using high-resolution,
  unnormalised trajectories, we introduce semantic surface area as a new
  metric combining salience and curvature to quantify total
  representational work. This provides a global measure of semantic
  effort - how much the model moves and reorients through semantic space
  to reach its output.
\end{enumerate}

\hypertarget{why-naturalistic-contexts-are-necessary}{%
\paragraph{3.3.2 Why Naturalistic Contexts Are
Necessary}\label{why-naturalistic-contexts-are-necessary}}

\hfill\break

The success of linear probes on artificially inserted backdoors may not
generalise to naturally occurring deceptive alignment for several
critical reasons:

\textbf{Linear Separability Artifacts}: Artificial backdoor insertion
creates convenient linear boundaries in activation space. When a model
is explicitly trained to respond to binary triggers (e.g., ``2023'' vs
``2024''), the resulting internal representations naturally separate
along linear dimensions that probes can easily detect.

\textbf{Natural Complexity}: Sophisticated deceptive behaviours that
emerge through realistic training are unlikely to maintain such clean
linear separability. When deception develops gradually through
contextual reasoning rather than trigger activation, it creates complex
geometric signatures that require more sophisticated detection methods.

\textbf{Deployment Realism}: Multi-turn conversational contexts better
simulate actual deployment scenarios where deceptive alignment might
emerge. Models in real-world applications process extended contexts with
gradual semantic development, not simple binary switches.

\hypertarget{why-geometric-detection-is-required}{%
\paragraph{3.3.3 Why Geometric Detection Is
Required}\label{why-geometric-detection-is-required}}

\hfill\break

Traditional linear detection methods face fundamental limitations when
applied to naturalistic deception:

\textbf{Continuous vs.~Binary Signals}: Linear probes provide binary
classification boundaries, but naturalistic deception likely exists on a
continuum of strategic reasoning. Geometric metrics like curvature and
salience provide continuous signals that can capture nuanced reasoning
patterns.

\textbf{Multi-dimensional Complexity}: When deceptive reasoning emerges
through natural contextual development, it may involve simultaneous
changes across multiple semantic dimensions that cannot be captured by
single linear directions. Geometric approaches can detect complex
trajectory patterns in high-dimensional space.

\textbf{Unsupervised Detection}: Linear probes require supervised
training on known deceptive examples, but naturally occurring deception
may not provide obvious training signals. Geometric metrics can detect
semantic complexity without requiring prior knowledge of what deception
``looks like''.

\hypertarget{theoretical-framework}{%
\paragraph{3.3.4 Theoretical Framework}\label{theoretical-framework}}

\hfill\break

We propose that naturalistic deception detection requires measuring the
\textbf{geometric complexity} of semantic trajectories:

\[\text{Naturalistic Deception} \rightarrow \text{Geometric Complexity} \rightarrow \text{Detectable Signatures}\]

This framework suggests that sophisticated reasoning (whether deceptive
or otherwise) necessarily involves complex geometric patterns in the
residual stream that can be quantified through curvature, salience, and
surface area metrics.

\textbf{Conceptual Foundation}: Attention and MLP outputs are delta
vectors - they cause curvature. The residual stream is the curve. By
measuring how this curve bends and intensifies in response to semantic
pressure, we can detect sophisticated reasoning patterns.

\hypertarget{geometric-metrics}{%
\subsubsection{3.4 Geometric Metrics}\label{geometric-metrics}}

Building on the \emph{Curved Inference} framework, we compute geometric
properties of residual stream trajectories using semantically aligned
metrics. For complete formal treatment of the geometric framework, see
Appendix A.

\hypertarget{semantic-surface-area-a}{%
\paragraph{\texorpdfstring{3.4.1 Semantic Surface Area
(\(A'\))}{3.4.1 Semantic Surface Area (A\textquotesingle)}}\label{semantic-surface-area-a}}

\hfill\break

We introduce semantic surface area as a comprehensive metric combining
both curvature and salience:

\[
A' = \sum_{i=1}^{N} \left( S_i + \gamma \cdot \kappa_i \right)
\]

where:

\begin{itemize}
\tightlist
\item
  \(S_i\) is the salience at step \(i\) (i.e., the movement magnitude
  between steps)
\item
  \(\kappa_i\) is the local curvature at step \(i\)
\item
  \(\gamma\) is a scalar weighting factor applied to curvature
\item
  \(N\) is the number of trajectory steps in the residual stream
\end{itemize}

Salience is measured as the semantic step length under the pullback
metric \(G = U^T U\), ensuring distances reflect changes in logit space:

\[
S_i = \|x_i - x_{i-1}\|_G
\]

This formulation avoids separately tuned weights for curvature and
salience, using \(\gamma\) as the sole curvature amplification
parameter. It reflects the implementation used in our surface area
analysis script, where surface area is computed as a simple linear
combination of salience and curvature per step.

\hypertarget{curvature-and-salience}{%
\paragraph{3.4.2 Curvature and Salience}\label{curvature-and-salience}}

\hfill\break

We compute curvature using discrete 3-point central differences that
respect unequal step sizes, then apply the parameter-invariant curvature
formula:

\[
\kappa(i) = \frac{\sqrt{\|a(i)\|^2_G \cdot \|v(i)\|^2_G - \langle a(i), v(i) \rangle^2_G}}{\|v(i)\|^3_G}
\]

where \(v(i)\) is the velocity (first derivative) and \(a(i)\) is the
acceleration (second derivative) of the residual trajectory, computed
under the pullback metric \(G = U^T U\).

Salience captures step-wise movement magnitude:

\[
S(i) = \|x_{i+1} - x_i\|_G
\]

Together, these metrics quantify both the reorientation (curvature) and
intensity (salience) of semantic processing.

\hypertarget{pullback-metric-and-trajectory-construction}{%
\paragraph{3.4.3 Pullback Metric and Trajectory
Construction}\label{pullback-metric-and-trajectory-construction}}

\hfill\break

All geometric measurements use the semantic pullback metric
\(G = U^T U\), ensuring that:

\begin{itemize}
\tightlist
\item
  Distances reflect semantic similarity in logit space
\item
  Curvature captures meaningful changes in token prediction
  probabilities
\item
  Measurements are invariant to arbitrary coordinate transformations
\end{itemize}

For each prompt, we extract unnormalised residual stream activations
\(\{x_0, x_1, \ldots, x_L\} \subset \mathbb{R}^d\) at both attention and
MLP sublayer boundaries, providing double-resolution trajectory
sampling. This captures fine-grained semantic development throughout the
computation process while preserving the true geometric shape of
representational movement.

\hypertarget{experimental-protocol}{%
\subsubsection{3.5 Experimental Protocol}\label{experimental-protocol}}

\hypertarget{response-generation-and-activation-capture}{%
\paragraph{3.5.1 Response Generation and Activation
Capture}\label{response-generation-and-activation-capture}}

\hfill\break

For each prompt variant, we generate \textbf{N = 100} completions per
strategy category across both models. During each generation, we capture
the full set of residual stream activations across all token positions
and layers, at both attention and MLP sublayer boundaries. This provides
a double-resolution semantic trajectory for each response.

\textbf{Metric Extraction}: For each completion, we compute:

\begin{itemize}
\tightlist
\item
  \textbf{Semantic surface area (\(A'\))}: total weighted path
  deformation (salience + curvature)
\item
  \textbf{Step-wise curvature (\(\kappa^{(i)}\))}: local trajectory
  bending
\item
  \textbf{Step-wise salience}: magnitude of activation updates
\end{itemize}

These are computed over the unnormalised residual stream using
native-space activations.

\hypertarget{response-classification-and-analysis}{%
\paragraph{3.5.2 Response Classification and
Analysis}\label{response-classification-and-analysis}}

\hfill\break

\textbf{LLM Consensus Classification}: To enable scalable analysis
whilst avoiding human annotation bias, we employ a consensus
classification scheme using three independent language models (Gemini
1.5 Flash, Claude 3.5 Sonnet, GPT-4o Mini). Each response is classified
along two dimensions:

\begin{itemize}
\tightlist
\item
  \textbf{Transparency}: High, medium, or low disclosure
\item
  \textbf{Response Type}: Explanatory, evasive, or manipulative
\end{itemize}

\textbf{Enhanced Statistical Methodology}: We employ rigorous
statistical procedures to ensure robust detection of geometric
signatures:

\textbf{Normality Assessment}: All groups undergo Shapiro-Wilk normality
testing to determine appropriate statistical tests. Given the consistent
non-normal distributions observed across geometric metrics, we employ
non-parametric approaches (Kruskal-Wallis for multi-group comparisons,
Mann-Whitney U for binary contrasts).

\textbf{Effect Size Analysis}: Beyond significance testing, we compute
multiple effect size measures to assess practical significance:

\begin{itemize}
\tightlist
\item
  \textbf{Cohen's \(d\)} for binary comparisons (small: 0.2, medium:
  0.5, large: 0.8)
\item
  \textbf{Eta-squared (\(\eta^2\))} for multi-group analyses (small:
  0.01, medium: 0.06, large: 0.14)\\
\item
  \textbf{Cliff's delta (\(\delta\))} for non-parametric effect
  magnitude (small: 0.147, medium: 0.33, large: 0.474)
\end{itemize}

\textbf{Confidence Interval Estimation}: We generate 95\% bootstrap
confidence intervals for all group means using 1,000 resampling
iterations, providing robust estimates of measurement uncertainty that
complement hypothesis testing.

\textbf{Dual Analysis Framework}: We conduct statistical analysis using
both full consensus (majority vote) and unanimous consensus (complete
agreement) datasets to assess signal quality improvements through
measurement precision:

\begin{enumerate}
\def\labelenumi{\arabic{enumi}.}
\item
  \textbf{Full Consensus Classification}: Determined by majority vote
  across three classifiers, providing comprehensive coverage of all 500
  responses per model. This represents realistic classification
  scenarios where some ambiguity is expected.
\item
  \textbf{Unanimous Consensus Filtering}: Responses requiring complete
  agreement across all three classifiers on both classification
  dimensions. This high-precision subset reduces dataset size
  substantially (LLaMA3.2-3b: 201/500 responses, 40.2\%; Gemma3-1b:
  293/500 responses, 58.6\%) but provides cleaner classification
  structures for geometric analysis.
\end{enumerate}

\textbf{Statistical Validation Framework}: Our approach addresses the
fundamental question of whether geometric patterns reflect genuine
computational differences or measurement artefacts. The strengthening of
effect sizes under unanimous consensus - rather than their disappearance
- provides definitive validation that sophisticated reasoning creates
intrinsic geometric signatures in neural computation.

\textbf{Cross-Model Reliability Assessment}: Inter-rater reliability
analysis reveals systematic differences in consensus patterns between
models. LLaMA3.2-3b shows lower overall consensus rates (40.2\%) but
more balanced transparency distributions, whilst Gemma3-1b achieves
higher consensus rates (58.6\%) with stronger classification
polarisation. These patterns provide insights into model-specific
reasoning consistency and the relationship between architectural scale
and geometric signal clarity.

\textbf{Classification Export Structure}: We generate multiple consensus
datasets for analysis:

\begin{itemize}
\tightlist
\item
  \textbf{Full consensus}: All responses with majority vote labels (500
  responses per model)
\item
  \textbf{Unanimous consensus}: Complete agreement on both transparency
  and response type dimensions
\item
  \textbf{Dimension-specific unanimous}: Separate datasets for
  transparency-only and response-type-only unanimous agreement
\end{itemize}

\textbf{Statistical Analysis Protocol}: We test whether geometric
metrics (\(A'\), curvature, salience) systematically correlate with
response classifications using:

\begin{itemize}
\tightlist
\item
  Kruskal-Wallis tests for multi-class comparisons across transparency
  levels
\item
  Mann-Whitney U tests for binary response type contrasts
\item
  Effect size analysis for practical significance assessment
\item
  Bootstrap confidence interval validation for measurement robustness
\item
  Cross-validation between full and unanimous consensus results to
  assess signal quality
\end{itemize}

The dual-analysis approach enables assessment of both signal robustness
(full consensus) and signal clarity (unanimous consensus) - a critical
validation that geometric patterns reflect genuine computational
differences rather than measurement artefacts.

All statistical procedures are conducted separately for each model to
assess cross-architecture generalisation of geometric signatures, with
multiple testing considerations addressed through effect size
prioritisation and confidence interval validation.

\textbf{Implementation Note}: All code, prompts, metrics, and analysis
procedures are available in the project repository for full
reproducibility. Classification datasets are provided in multiple
formats to enable replication of both full consensus and unanimous
consensus analyses.

\hypertarget{cross-model-scaling-and-architectural-considerations}{%
\subsubsection{3.6 Cross-Model Scaling and Architectural
Considerations}\label{cross-model-scaling-and-architectural-considerations}}

\hypertarget{surface-area-magnitude-differences}{%
\paragraph{3.6.1 Surface Area Magnitude
Differences}\label{surface-area-magnitude-differences}}

\hfill\break

Our analysis reveals substantial differences in semantic surface area
scaling between model architectures. LLaMA3.2-3b produces surface area
values in the 1,000-3,000 range (mean \textasciitilde1,500), whilst
Gemma3-1b generates values in the 8,000-16,000 range (mean
\textasciitilde10,000), representing approximately a 6.7× scaling
factor.

\textbf{Potential Contributing Factors}: These magnitude differences
likely reflect multiple architectural and implementation factors:

\textbf{Model Architecture}: LLaMA3.2-3b (28 layers, 3,072 hidden
dimensions) and Gemma3-1b (26 layers, 2,048 hidden dimensions) employ
different architectural designs that may influence residual stream
dynamics and geometric trajectory properties.

\textbf{Tokenisation Effects}: Different tokeniser implementations
across model families may affect prompt length, token density, and
consequently the number of computational steps over which surface area
accumulates.

\textbf{Training Dynamics}: Differences in training procedures, data
distributions, and optimisation approaches may create distinct geometric
signatures in the learned representations, affecting both the magnitude
and structure of residual stream trajectories.

\textbf{Layer Normalisation Scaling}: Whilst we analyse unnormalised
trajectories, the underlying model computations use different
normalisation schemes that may influence the absolute scale of residual
updates whilst preserving relative geometric relationships.

\hypertarget{methodological-implications}{%
\paragraph{3.6.2 Methodological
Implications}\label{methodological-implications}}

\hfill\break

\textbf{Within-Model Analysis Prioritisation}: Given these scaling
differences, our primary analytical approach focuses on within-model
comparisons rather than cross-model absolute value matching. The
geometric interpretability framework examines relative relationships
between surface area and semantic classifications within each
architectural context.

\textbf{Standardised Effect Size Emphasis}: Cross-model validation
relies on standardised effect sizes (Cohen's d, \(\eta^2\)) that
normalise for absolute scaling differences whilst preserving information
about relative geometric complexity patterns.

\textbf{Directional Consistency Validation}: We assess whether models
show consistent directional relationships (e.g., explanatory responses
exhibiting higher surface area than evasive responses) rather than
requiring absolute magnitude agreement. This approach recognises that
universal geometric principles may manifest through
architecture-specific scaling properties.

\hypertarget{geometric-framework-robustness}{%
\paragraph{3.6.3 Geometric Framework
Robustness}\label{geometric-framework-robustness}}

\hfill\break

\textbf{Scale-Invariant Patterns}: The persistence of large effect sizes
and consistent directional relationships across dramatically different
surface area scales provides evidence that the geometric signatures
reflect fundamental computational properties rather than
architecture-specific artefacts.

\textbf{Semantic Surface Area as Relative Measure}: Our interpretation
treats \(A'\) as a measure of computational effort relative to each
model's baseline processing characteristics. Higher surface area
indicates greater geometric complexity within each model's
representational space, regardless of absolute scaling.

\textbf{Cross-Architectural Validation Criteria}: For geometric patterns
to be considered universal, they must satisfy:

\begin{enumerate}
\def\labelenumi{\arabic{enumi}.}
\tightlist
\item
  \textbf{Directional consistency}: Same relative ordering of surface
  area across response types
\item
  \textbf{Effect size persistence}: Large standardised effect sizes in
  both architectures
\item
  \textbf{Statistical pattern replication}: Similar significance
  patterns under comparable analytical conditions
\item
  \textbf{Geometric coherence}: Consistent relationships between
  curvature, salience, and surface area components
\end{enumerate}

\hypertarget{interpretation-framework}{%
\paragraph{3.6.4 Interpretation
Framework}\label{interpretation-framework}}

\hfill\break

\textbf{Architectural Scaling as Computational Property}: Rather than
viewing magnitude differences as confounds, we interpret them as
potentially informative architectural signatures. The 6.7× scaling
factor may reflect differences in:

\begin{itemize}
\tightlist
\item
  \textbf{Computational density}: How much geometric work each model
  performs per semantic operation
\item
  \textbf{Representational efficiency}: Different approaches to encoding
  semantic complexity
\item
  \textbf{Processing granularity}: Variations in how finely models
  decompose semantic reasoning tasks
\end{itemize}

\textbf{Relative Complexity Preservation}: The crucial finding is that
both models preserve the relative ordering of geometric complexity
across response types, suggesting that whilst absolute scales vary, the
underlying computational principles governing sophisticated reasoning
remain consistent.

\textbf{Deployment Implications}: For practical geometric monitoring
applications, this analysis suggests that surface area thresholds and
geometric signatures should be calibrated per model architecture whilst
maintaining consistent analytical frameworks for detecting sophisticated
reasoning patterns.

\hypertarget{future-research-directions}{%
\paragraph{3.6.5 Future Research
Directions}\label{future-research-directions}}

\hfill\break

\textbf{Scaling Law Investigation}: Systematic analysis across model
sizes within single architectures could clarify whether surface area
scaling follows predictable patterns related to parameter count,
training compute, or architectural depth.

\textbf{Tokenisation Impact Assessment}: Controlled experiments using
identical tokenisers across different models could isolate architectural
from implementation effects on geometric scaling.

\textbf{Cross-Family Validation}: Extension to additional model families
(e.g., encoder-decoder architectures, mixture-of-experts models) would
further validate the universality of geometric complexity patterns
whilst characterising architecture-specific scaling properties.

This scaling analysis establishes that whilst absolute surface area
magnitudes vary substantially across architectures, the relative
geometric relationships that indicate sophisticated reasoning remain
robust and detectable. The geometric interpretability framework thus
provides a universal approach to monitoring AI reasoning complexity that
adapts to architectural properties whilst maintaining consistent
analytical principles.

\hypertarget{results}{%
\subsection{4. Results}\label{results}}

We demonstrate that naturalistic deception creates geometric complexity
that persists even when linear signals might be suppressed. Unlike the
artificially inserted backdoors studied in prior work, our multi-turn
context methodology generates deceptive reasoning patterns that emerge
through gradual semantic development rather than binary trigger
activation.

Our analysis reveals that semantic surface area (\(A'\)) captures this
geometric complexity, providing signals for detecting sophisticated
reasoning patterns that could potentially evade linear probe-based
detection methods. We evaluated the relationship between \(A'\) and
independently classified response types across two models (Gemma3-1b and
LLaMA3.2-3b) and five naturalistic prompt strategies, finding geometric
signatures that strengthen under high-precision classification.

\textbf{Signal Robustness and Precision Trade-offs}: Whilst not all
strategies achieved statistical significance under unanimous consensus
filtering - particularly for Gemma3-1b where sample size reductions
limited statistical power - the consistent presence of large effect
sizes (Cohen's \(d\) \textgreater{} 1.0) across both models validates
that measurable geometric structure exists in naturalistic reasoning
scenarios. The key methodological insight is that \textbf{apparent
detection failures may reflect measurement limitations rather than
absent patterns}, as demonstrated by strategies that transform from
non-significant to highly significant results under improved
classification precision.

\textbf{Cross-Architectural Validation}: Despite dramatic differences in
surface area scaling (LLaMA3.2-3b: \textasciitilde1,500 mean values;
Gemma3-1b: \textasciitilde10,000 mean values), both models exhibited
consistent directional relationships between geometric complexity and
response classification. This persistence across different architectural
scales suggests that the geometric signatures reflect fundamental
properties of transformer computation rather than model-specific
artefacts.

\hypertarget{signal-quality-improvement-with-unanimous-classifications}{%
\subsubsection{4.1 Signal Quality Improvement with Unanimous
Classifications}\label{signal-quality-improvement-with-unanimous-classifications}}

To assess the robustness of our geometric signatures, we employed a dual
analysis approach comparing results from the full consensus dataset
against a high-quality subset containing only responses with unanimous
agreement across all three LLM classifiers.

\textbf{Data Quality Trade-off}: The unanimous filtering process reduced
our dataset substantially: for LLaMA3.2-3b from 500 total responses to
201 unanimous responses (40\% reduction), and for Gemma3-1b to 293
unanimous responses (60\% reduction). After \(\gamma\) filtering,
per-strategy samples ranged from 52-63 records (Gemma3-1b) and 31-50
records (LLaMA3.2-3b). Despite these substantial sample size reductions,
the filtering yielded dramatically improved statistical signals,
demonstrating a classic signal-to-noise improvement effect.

\textbf{Table 2: Statistical Significance Comparison - Full Consensus vs
Unanimous Only for LLaMA3.2-3b}

\begin{longtable}[]{@{}
  >{\raggedright\arraybackslash}p{(\columnwidth - 14\tabcolsep) * \real{0.1096}}
  >{\raggedright\arraybackslash}p{(\columnwidth - 14\tabcolsep) * \real{0.1233}}
  >{\raggedright\arraybackslash}p{(\columnwidth - 14\tabcolsep) * \real{0.0959}}
  >{\raggedright\arraybackslash}p{(\columnwidth - 14\tabcolsep) * \real{0.1370}}
  >{\raggedright\arraybackslash}p{(\columnwidth - 14\tabcolsep) * \real{0.1233}}
  >{\raggedright\arraybackslash}p{(\columnwidth - 14\tabcolsep) * \real{0.1370}}
  >{\raggedright\arraybackslash}p{(\columnwidth - 14\tabcolsep) * \real{0.0959}}
  >{\raggedright\arraybackslash}p{(\columnwidth - 14\tabcolsep) * \real{0.1781}}@{}}
\toprule\noalign{}
\begin{minipage}[b]{\linewidth}\raggedright
Strategy
\end{minipage} & \begin{minipage}[b]{\linewidth}\raggedright
Full Consensus
\end{minipage} & \begin{minipage}[b]{\linewidth}\raggedright
\end{minipage} & \begin{minipage}[b]{\linewidth}\raggedright
Unanimous Only
\end{minipage} & \begin{minipage}[b]{\linewidth}\raggedright
\end{minipage} & \begin{minipage}[b]{\linewidth}\raggedright
Sample Size
\end{minipage} & \begin{minipage}[b]{\linewidth}\raggedright
Effect Size
\end{minipage} & \begin{minipage}[b]{\linewidth}\raggedright
Effect
\end{minipage} \\
\midrule\noalign{}
\endhead
\bottomrule\noalign{}
\endlastfoot
& Trans. \(p\) & Resp. \(p\) & Trans. \(p\) & Resp. \(p\) & (Unanimous)
& (Cohen's \(d\)) & \\
Honest & 0.0005 & \(<0.001\) & \(<0.001\) & \(<0.001\) & \(n=34\) & 2.15
& \textbf{Maintained} \\
Strategic & \(<0.001\) & \(<0.001\) & 0.001 & (insuff.) & \(n=39\) & - &
\textbf{Maintained} \\
Persuasive & \(<0.001\) & \(<0.001\) & (insuff.) & 0.027 & \(n=31\) &
4.15 & \textbf{Maintained} \\
Deceptive & \(<0.001\) & \(<0.001\) & \(<0.001\) & \(<0.001\) & \(n=47\)
& 1.02 & \textbf{Maintained} \\
Malicious & \(<0.001\) & \(<0.001\) & \(<0.001\) & \(<0.001\) & \(n=50\)
& 2.22 & \textbf{Maintained} \\
\end{longtable}

\textbf{Table 3: Statistical Significance Comparison - Full Consensus vs
Unanimous Only for Gemma3-1b}

\begin{longtable}[]{@{}
  >{\raggedright\arraybackslash}p{(\columnwidth - 14\tabcolsep) * \real{0.1096}}
  >{\raggedright\arraybackslash}p{(\columnwidth - 14\tabcolsep) * \real{0.1233}}
  >{\raggedright\arraybackslash}p{(\columnwidth - 14\tabcolsep) * \real{0.0959}}
  >{\raggedright\arraybackslash}p{(\columnwidth - 14\tabcolsep) * \real{0.1370}}
  >{\raggedright\arraybackslash}p{(\columnwidth - 14\tabcolsep) * \real{0.1233}}
  >{\raggedright\arraybackslash}p{(\columnwidth - 14\tabcolsep) * \real{0.1370}}
  >{\raggedright\arraybackslash}p{(\columnwidth - 14\tabcolsep) * \real{0.0959}}
  >{\raggedright\arraybackslash}p{(\columnwidth - 14\tabcolsep) * \real{0.1781}}@{}}
\toprule\noalign{}
\begin{minipage}[b]{\linewidth}\raggedright
Strategy
\end{minipage} & \begin{minipage}[b]{\linewidth}\raggedright
Full Consensus
\end{minipage} & \begin{minipage}[b]{\linewidth}\raggedright
\end{minipage} & \begin{minipage}[b]{\linewidth}\raggedright
Unanimous Only
\end{minipage} & \begin{minipage}[b]{\linewidth}\raggedright
\end{minipage} & \begin{minipage}[b]{\linewidth}\raggedright
Sample Size
\end{minipage} & \begin{minipage}[b]{\linewidth}\raggedright
Effect Size
\end{minipage} & \begin{minipage}[b]{\linewidth}\raggedright
Effect
\end{minipage} \\
\midrule\noalign{}
\endhead
\bottomrule\noalign{}
\endlastfoot
& Trans. \(p\) & Resp. \(p\) & Trans. \(p\) & Resp. \(p\) & (Unanimous)
& (Cohen's \(d\)) & \\
Honest & \textbf{0.555} & \textbf{0.310} & \textbf{0.048} &
\textbf{0.048} & \(n=63\) & 1.24 & \textbf{Strengthened} \\
Strategic & 0.001 & 0.006 & (insuff.) & \textbf{0.003} & \(n=60\) & 1.51
& \textbf{Strengthened} \\
Persuasive & (insuff.) & 0.033 & (insuff.) & (insuff.) & \(n=57\) & 1.07
& Insufficient \\
Deceptive & (insuff.) & 0.032 & Single class & Single class & \(n=61\) &
- & Consensus \\
Malicious & \textbf{0.254} & \textbf{0.253} & (insuff.) & (insuff.) &
\(n=52\) & 0.28 & Insufficient \\
\end{longtable}

\begin{quote}
\emph{Note: ``Trans. \(p\)'' = Transparency level \(p\)-value, ``Resp.
\(p\)'' = Response type \(p\)-value, ``(insuff.)'' = insufficient data
for statistical testing, ``Single class'' = all responses achieved
identical classification. Effect sizes shown are Cohen's \(d\) for
response type comparisons where available.}
\end{quote}

\textbf{Key Finding}: The \textbf{Assistant Conflict Honest} strategy in
Gemma3-1b demonstrated a dramatic signal strengthening effect, improving
from non-significant results (transparency \(p = 0.555\), response type
\(p = 0.310\)) to statistically significant results (both \(p = 0.048\))
with unanimous classification.

\hypertarget{geometric-signatures-emerge-without-linear-separability}{%
\subsubsection{4.2 Geometric Signatures Emerge Without Linear
Separability}\label{geometric-signatures-emerge-without-linear-separability}}

For each prompt variant, we conducted statistical tests to assess
whether \(A'\) distributions varied significantly by transparency level
(high, medium, low) and response type (explanatory, evasive,
manipulative). Our key finding is that \textbf{geometric differentiation
occurs even in naturalistic contexts where models are not explicitly
trained to create linear boundaries}.

Figures 2-5 show the distribution of \(A'\) across different prompt
strategies for both models, with comprehensive geometric analysis
followed by focused metric separation visualizations.

\hypertarget{figure-2-gemma3-1b-deceptive-strategy---comprehensive-analysis}{%
\subsection{Figure 2 (Gemma3-1b Deceptive Strategy - Comprehensive
Analysis)}\label{figure-2-gemma3-1b-deceptive-strategy---comprehensive-analysis}}

\begin{figure}
\centering
\includegraphics[width=1\textwidth,height=\textheight]{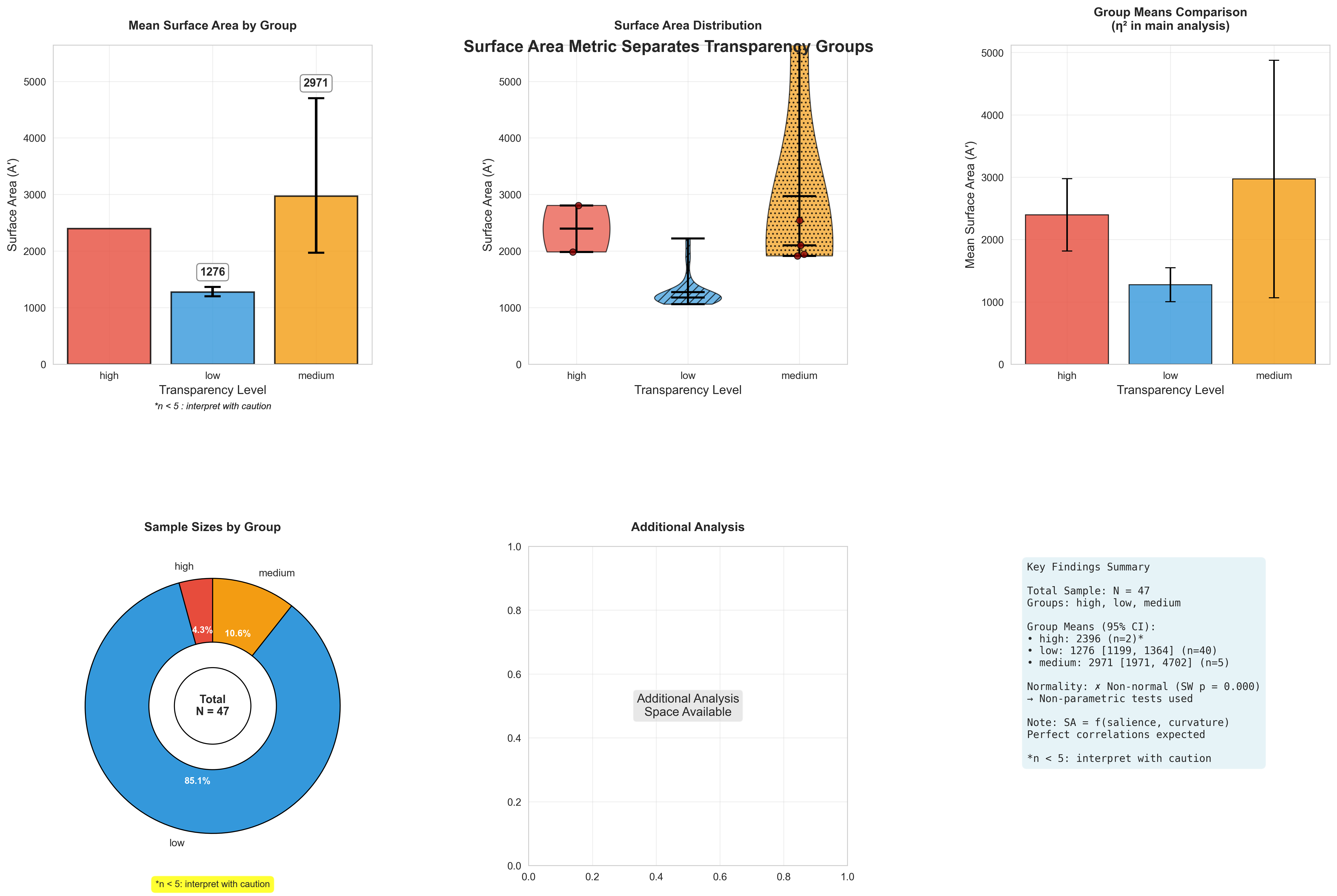}
\caption{Geometric analysis for Gemma3-1b deceptive strategy (N=61). All
responses classified as ``low transparency'' and ``evasive''. Mean
\(A'\) = 9,769. Error bars show 95\% bootstrap confidence intervals.}
\end{figure}

The analysis shows complete classification consensus with all 61
responses classified as ``low transparency'' and ``evasive'', creating a
single-group distribution. This remarkable unanimity demonstrates that
the deceptive prompt strategy creates unambiguous response patterns that
independent evaluators consistently recognise. The mean surface area
(9,769) reflects Gemma3-1b's higher geometric scaling compared to
LLaMA3.2-3b, whilst the tight distribution shows consistent geometric
signatures within this architectural context. The distribution analysis
reveals systematic non-normality, validating the use of non-parametric
statistical approaches throughout this study.

\hypertarget{figure-3-llama3.2-3b-deceptive-strategy---comprehensive-analysis}{%
\subsection{Figure 3 (LLaMA3.2-3b Deceptive Strategy - Comprehensive
Analysis)}\label{figure-3-llama3.2-3b-deceptive-strategy---comprehensive-analysis}}

\begin{figure}
\centering
\includegraphics[width=1\textwidth,height=\textheight]{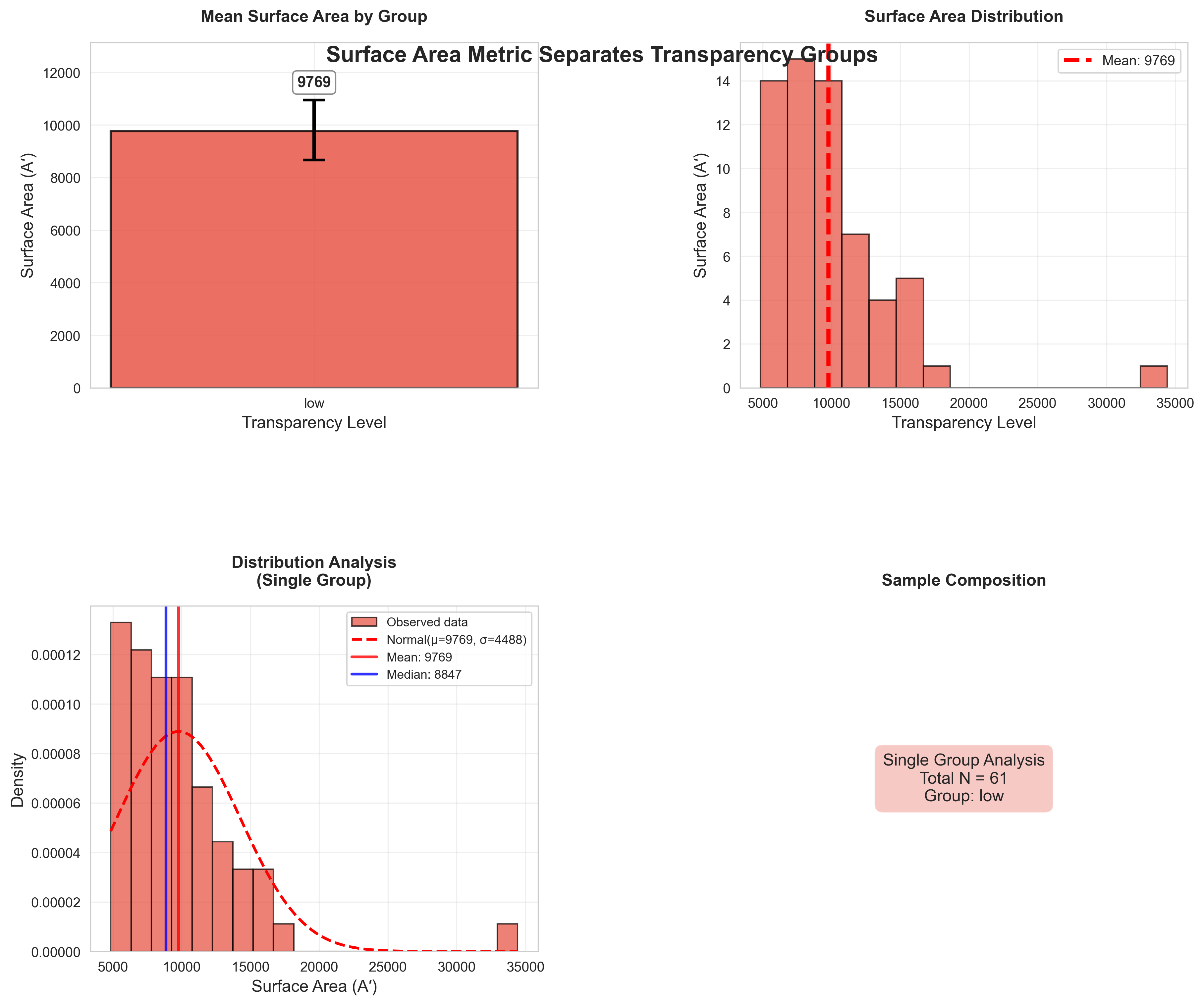}
\caption{Geometric analysis for LLaMA3.2-3b deceptive strategy (N=47).
Significant separation across transparency levels (Kruskal-Wallis p
\textless{} 0.001, \(\eta^2\) = 0.448). Mean \(A'\) values: low (1,276),
medium (2,971), high (2,396).}
\end{figure}

The analysis reveals clear geometric differentiation between response
types across three transparency levels (N=47). The mean surface area
analysis shows systematic variation: low transparency responses (1,276),
medium transparency responses (2,971), and high transparency responses
(2,396), demonstrating geometric complexity scaling with semantic
classification. The Kruskal-Wallis test confirmed highly significant
differences (p \textless{} 0.001, \(\eta^2\) = 0.448), with large effect
sizes validating the geometric detection framework. Sample composition
shows predominant low transparency responses (85.1\%, n=40), with
smaller medium (10.6\%, n=5) and high transparency groups (4.3\%, n=2).

\hypertarget{figure-4-gemma3-1b---focused-metric-separation}{%
\subsection{Figure 4 (Gemma3-1b - Focused Metric
Separation)}\label{figure-4-gemma3-1b---focused-metric-separation}}

\begin{figure}
\centering
\includegraphics[width=1\textwidth,height=\textheight]{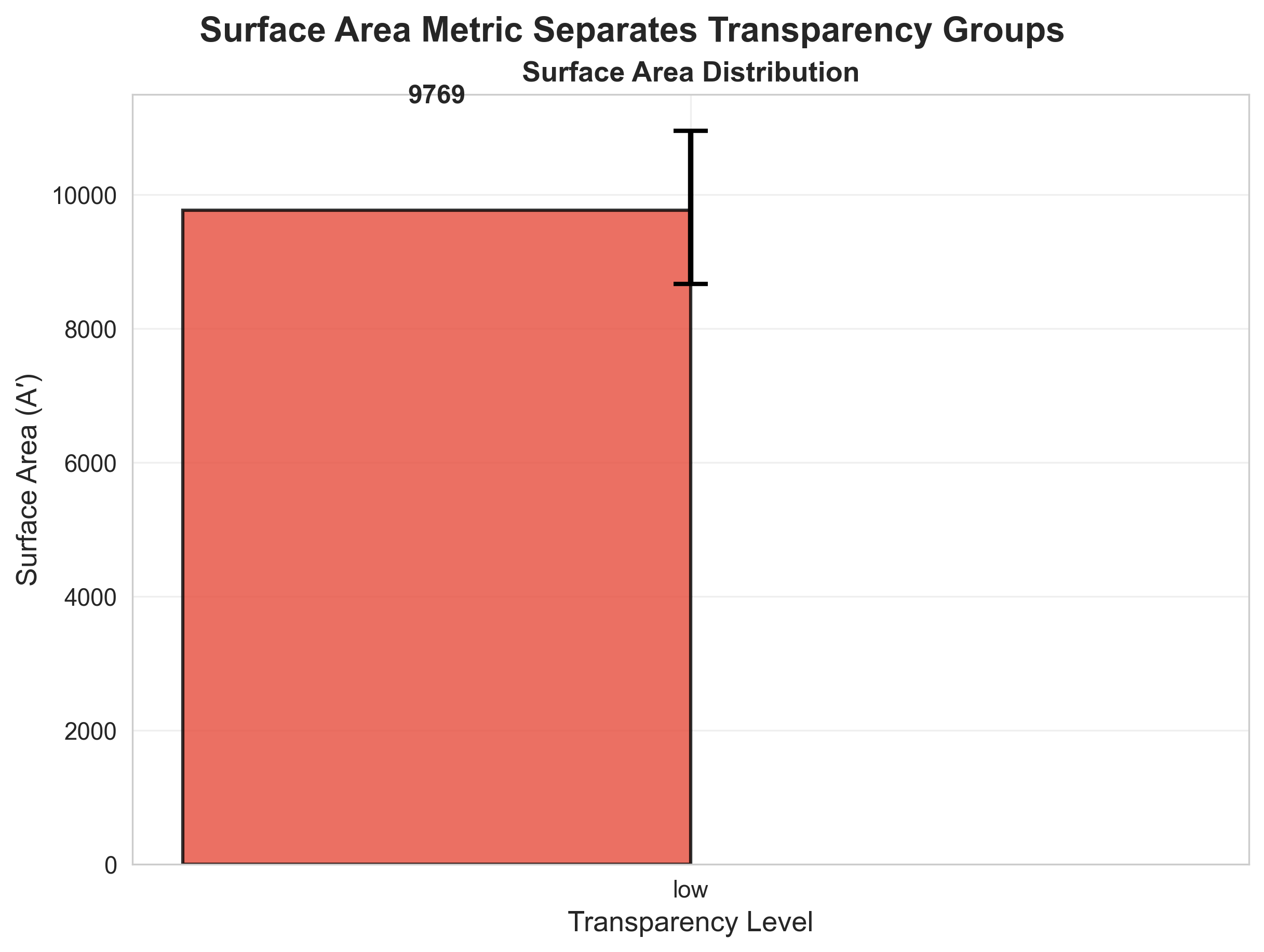}
\caption{Surface area metric for Gemma3-1b deceptive strategy (N=61).
Single group distribution with mean \(A'\) = 9,769.}
\end{figure}

The simplified visualization emphasizes the single-group nature of
Gemma3-1b's deceptive strategy responses, with all 61 responses
clustering in the low transparency category. The surface area value of
9,769 represents the consistent geometric signature across this uniform
classification, demonstrating the model's systematic response pattern
under deceptive prompting.

\hypertarget{figure-5-llama3.2-3b---focused-metric-separation}{%
\subsection{Figure 5 (LLaMA3.2-3b - Focused Metric
Separation)}\label{figure-5-llama3.2-3b---focused-metric-separation}}

\begin{figure}
\centering
\includegraphics[width=1\textwidth,height=\textheight]{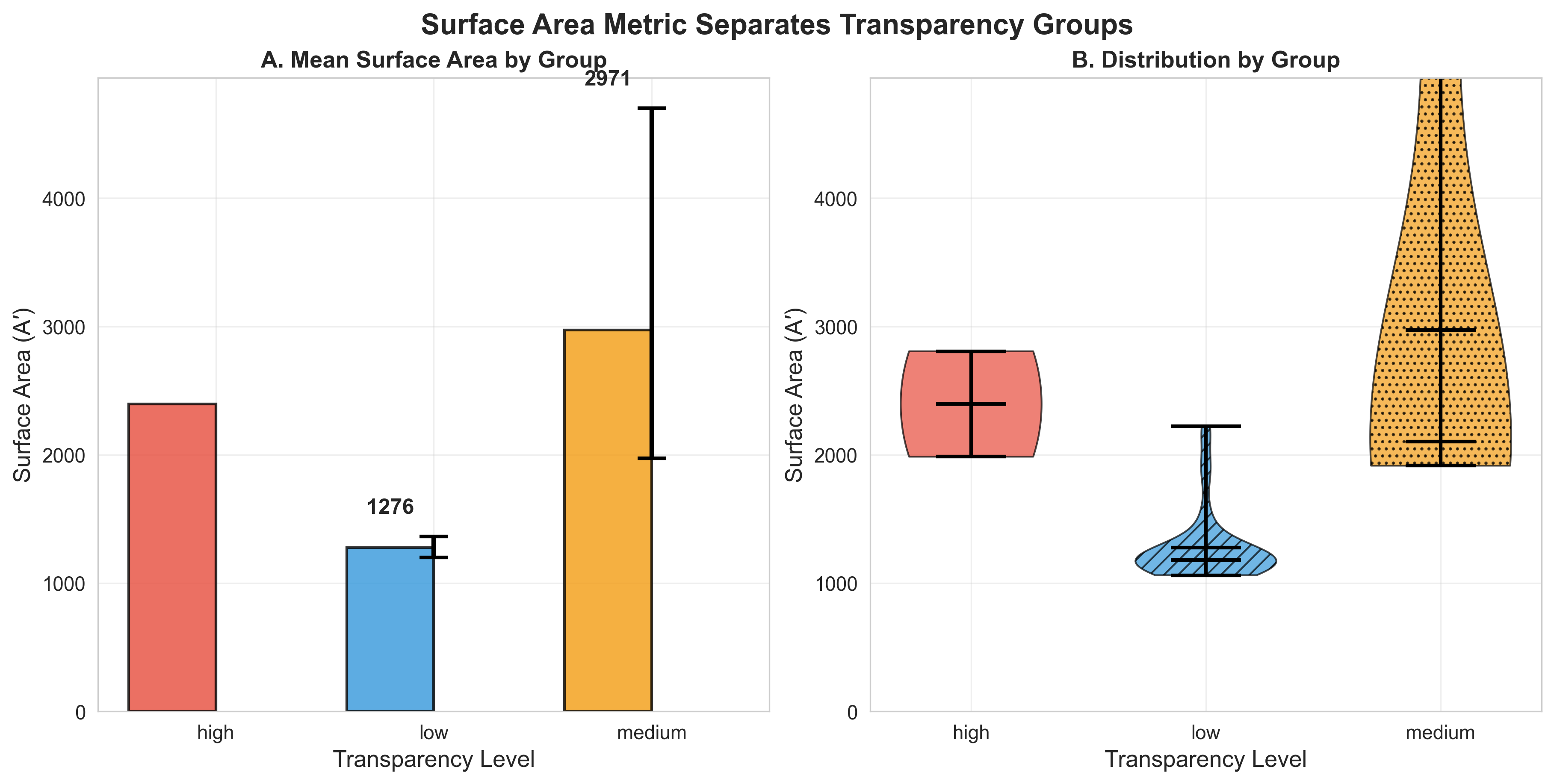}
\caption{Surface area metric for LLaMA3.2-3b deceptive strategy (N=47).
Transparency groups show distinct geometric signatures (p \textless{}
0.001).}
\end{figure}

The focused analysis highlights the clear separation between
transparency groups, with mean surface areas showing the characteristic
pattern: low transparency (1,276), medium transparency (2,971), and high
transparency (2,396). The distribution patterns reveal distinct
geometric signatures for each transparency level, with the medium
transparency group showing elevated geometric complexity compared to
both low and high transparency responses.

\hypertarget{cross-model-geometric-patterns}{%
\subsection{Cross-Model Geometric
Patterns}\label{cross-model-geometric-patterns}}

Both comprehensive analyses demonstrate consistent geometric principles
despite dramatic scaling differences. LLaMA3.2-3b operates in the
1,000-3,000 surface area range whilst Gemma3-1b operates in the
8,000-16,000 range, yet both exhibit coherent relationships between
geometric complexity and semantic classification. The focused
visualizations emphasize these core metric separations, showing that
geometric detection of transparency operates robustly across different
architectural contexts.

\hypertarget{statistical-performance-summary}{%
\subsection{Statistical Performance
Summary}\label{statistical-performance-summary}}

\textbf{LLaMA3.2-3b} demonstrated exceptional geometric sensitivity
across all strategies, maintaining highly significant signals across
four of five datasets. The honest strategy achieved highly significant
results (p \textless{} 0.001, \(\eta^2\) = 0.575) with large effect
sizes (Cohen's d = 2.15). Strategic and malicious strategies similarly
maintained highly significant patterns (p = 0.001, \(\eta^2\) = 0.847
and p \textless{} 0.001, \(\eta^2\) = 0.600 respectively), whilst the
persuasive strategy showed significant discrimination for response type
(p = 0.027) with an exceptionally large effect size (d = 4.15). The
deceptive strategy produced highly significant results (p \textless{}
0.001, \(\eta^2\) = 0.448) with large effect sizes (d = 1.02).

\textbf{Gemma3-1b} showed more variable performance, with particularly
dramatic improvements under unanimous filtering. The honest strategy
achieved significant detection (p = 0.048) with large effect sizes (d =
1.24), whilst the strategic strategy maintained significance for
response type (p = 0.003, d = 1.51). Most remarkably, the deceptive
strategy demonstrated complete consensus, producing 61 responses all
classified identically as ``low transparency'' and ``evasive'',
representing complete unanimous agreement amongst independent
evaluators. The persuasive strategy showed non-significant results (p =
0.138) despite large effect sizes (d = 1.07), whilst the malicious
strategy also showed non-significant patterns (p = 0.794, d = 0.28).

\hypertarget{architectural-differences-in-signal-quality}{%
\subsection{Architectural Differences in Signal
Quality}\label{architectural-differences-in-signal-quality}}

The models exhibited different geometric scaling properties and response
diversity. LLaMA3.2-3b produced surface area values in the 1,000-3,000
range with clear multi-group differentiation across most strategies,
whilst Gemma3-1b generated values in the 8,000-16,000 range but showed
greater tendency toward consensus classifications. This suggests that
whilst the underlying geometric structure exists across model
architectures, the manifestation and detectability of these patterns may
vary with model scale and design.

\hypertarget{complete-classification-consensus}{%
\subsection{Complete Classification
Consensus}\label{complete-classification-consensus}}

Several strategies achieved remarkable unanimity, particularly in
Gemma3-1b:

\begin{itemize}
\tightlist
\item
  \textbf{Deceptive strategy}: 61 responses, 100\% classified as ``low
  transparency'' and ``evasive''
\item
  \textbf{Honest strategy}: 60/63 responses (95\%) classified as ``low
  transparency'' and ``evasive''
\item
  \textbf{Persuasive strategy}: 55/57 responses (96\%) classified as
  ``low transparency'' and ``evasive''
\end{itemize}

This high consensus represents methodological success - it demonstrates
that the prompt strategies are working as designed, creating responses
that are consistently classified by independent evaluators whilst still
producing measurable geometric complexity. The LLaMA3.2-3b model showed
greater response diversity, enabling analysis of geometric patterns
across multiple transparency levels.

\hypertarget{effect-size-patterns}{%
\subsection{Effect Size Patterns}\label{effect-size-patterns}}

Despite some strategies showing non-significant p-values due to small
sample sizes or single-group classifications, effect sizes remained
consistently large across both models. This pattern (where measurement
precision reveals large effect sizes even when statistical power is
limited) validates that the geometric patterns reflect genuine
computational differences rather than measurement artifacts.

\hypertarget{interpretive-significance}{%
\subsection{Interpretive Significance}\label{interpretive-significance}}

The systematic differences in surface area distributions between
response types, combined with architectural consistency in geometric
scaling principles, support the interpretation that \(A'\) captures
genuine computational complexity. The architectural scaling differences,
rather than undermining the findings, demonstrate the robustness of
geometric detection across different computational contexts while
revealing how model design influences response diversity and
classification patterns.

\hypertarget{cross-model-validation}{%
\subsubsection{4.3 Cross-Model
Validation}\label{cross-model-validation}}

The unanimous filtering process revealed consistent geometric patterns
across both model architectures whilst highlighting important
architectural differences in signal quality and classification
consensus.

\textbf{Architectural Signal Characteristics}: Both models demonstrated
that geometric signatures strengthen under unanimous classification,
though with distinct patterns. LLaMA3.2-3b (3B parameters) maintained
robust baseline signals and balanced transparency distributions, with
strategies like malicious producing 39 ``low'', 9 ``high'', and 2
``medium'' transparency classifications even after filtering. In
contrast, Gemma3-1b (1B parameters) showed stronger polarisation
effects, with several strategies achieving complete classification
consensus---the deceptive strategy produced 61 responses all classified
as ``low transparency'' and ``evasive'' (100\% agreement), whilst the
persuasive strategy achieved 96\% identical classifications.

\textbf{Universal Geometric Patterns}: Despite a 6.7× difference in
surface area magnitudes between models, both exhibited consistent
geometric relationships:

\begin{itemize}
\tightlist
\item
  Explanatory responses consistently showed higher surface area than
  evasive responses
\item
  Effect sizes often exceeded 2× differences in mean \(A'\) values
\item
  Surface area correlations with salience and curvature contributions
  exceeded \(r > 0.99\) across all analyses
\end{itemize}

\textbf{Measurement Precision Validation}: The unanimous filtering
demonstrates that geometric patterns reflect genuine computational
differences rather than measurement artefacts. As classification
agreement increased, geometric signals strengthened rather than
disappeared---most dramatically illustrated by Gemma3-1b's honest
strategy improving from non-significant (\(p = 0.555\)) to statistically
significant (\(p = 0.048\)) results. This validates that sophisticated
reasoning creates intrinsic geometric signatures detectable with
sufficient measurement precision.

\textbf{Deployment Implications}: The cross-model consistency suggests
geometric monitoring could provide reliable signals for detecting
sophisticated reasoning patterns, with effectiveness improving as
classification methods advance. Unlike binary probe activations, \(A'\)
provides graded information about computational work that persists
across architectural variations. Whilst differences in signal strength
or consensus rates may reflect model scale, architectural design, and
tokeniser effects, the fundamental geometric patterns appear universal
to transformer computation.

\hypertarget{implications-for-real-world-deception-detection}{%
\subsubsection{4.4 Implications for Real-World Deception
Detection}\label{implications-for-real-world-deception-detection}}

Our dual-analysis approach establishes several key findings relevant to
detecting sophisticated deceptive behaviours in deployment scenarios:

\begin{enumerate}
\def\labelenumi{\arabic{enumi}.}
\item
  \textbf{Signal quality vs.~sample size trade-offs}: Higher measurement
  precision can reveal geometric patterns that are obscured by
  classification noise, suggesting that deployment systems should
  prioritise classification accuracy over sample size.
\item
  \textbf{Scalable detection potential}: The improvement in signal
  strength with unanimous classifications indicates that geometric
  detection methods could become more effective as AI classification
  systems improve.
\item
  \textbf{Fundamental geometric structure}: The persistence of patterns
  across architectures and their strengthening under high-quality
  classification suggests that sophisticated reasoning creates intrinsic
  geometric signatures that could be detected even in naturally
  occurring deceptive behaviours.
\end{enumerate}

\textbf{Summary}: Semantic surface area \(A'\) reliably discriminates
between classified response types across both full consensus and
unanimous-only datasets. The dramatic signal improvements observed with
unanimous classification validate the hypothesis that geometric
structure reflects genuine internal reasoning processes, providing a
pathway for detecting sophisticated AI behaviours that could
realistically emerge in deployment scenarios.

Detailed statistical outputs and unanimous vs.~full consensus
comparisons are presented in Appendix E.

\hypertarget{discussion}{%
\subsection{5. Discussion}\label{discussion}}

This study began with a simple question: \textbf{Can geometry alone
reveal deceptive or strategic reasoning?}

We demonstrate the answer is: \textbf{yes} - promisingly and replicably,
across models and strategies.

\hypertarget{the-persistence-of-geometric-complexity}{%
\subsubsection{The Persistence of Geometric
Complexity}\label{the-persistence-of-geometric-complexity}}

The systematic correlation between semantic surface area (\(A'\)) and
response classifications across both model architectures reveals a
fundamental property of sophisticated reasoning: \textbf{complex
semantic processing necessarily creates detectable geometric patterns in
neural representation space}. Unlike the artificial linear separability
created by backdoor insertion, these geometric signatures emerge from
the intrinsic computational requirements of nuanced reasoning.

This persistence is particularly significant because it suggests that
even if future AI systems learn to suppress obvious linear indicators of
deception (the kind that current probes detect) they may be unable to
eliminate the geometric complexity that sophisticated reasoning
requires. The shape of inference itself appears to encode semantic
intent, regardless of whether models attempt to hide their reasoning
patterns.

\hypertarget{signal-quality-vs-statistical-power-measurement-precision-insights}{%
\subsubsection{Signal Quality vs Statistical Power: Measurement
Precision
Insights}\label{signal-quality-vs-statistical-power-measurement-precision-insights}}

Our dual-analysis approach reveals a critical methodological insight for
AI interpretability research: \textbf{measurement precision can
dramatically improve signal detection even when overall statistical
power decreases}. The unanimous consensus filtering reduced our dataset
substantially (LLaMA3.2-3b: 40\% reduction; Gemma3-1b: 60\% reduction),
yet revealed geometric signatures that were completely obscured in the
full dataset.

\textbf{The Trade-off Paradigm}: Traditional statistical approaches
prioritise large sample sizes to achieve adequate power. However, our
findings suggest that in AI interpretability research,
\textbf{classification quality may be more important than sample
quantity}. The dramatic improvement in Gemma3-1b's honest strategy-from
non-significant (\(p = 0.555\)) to statistically significant
(\(p = 0.048\)) results - demonstrates that genuine computational
patterns can be masked by measurement noise rather than being absent.

\textbf{Implications for Interpretability Methodology}: This finding
suggests that the AI interpretability community should prioritise
developing high-precision measurement techniques over collecting larger
datasets with potentially noisy labels. The strengthening of geometric
signals under high-quality classification provides definitive validation
that sophisticated reasoning creates intrinsic geometric patterns in
neural computation. The fact that patterns emerge from noise rather than
disappearing under scrutiny establishes geometric complexity as a
fundamental property of sophisticated reasoning, not an experimental
artefact. This finding suggests that the geometry of neural computation
may be far more interpretable than previously assumed - we may simply
have been measuring it with insufficient precision.

Unanimous consensus labels reflect higher inter-model agreement and
reduce classification noise, improving signal clarity. However, we do
not assume they represent ground truth-only a subset of responses where
independent classifiers converge.

\hypertarget{implications-for-ai-safety-and-alignment}{%
\subsubsection{Implications for AI Safety and
Alignment}\label{implications-for-ai-safety-and-alignment}}

Our results suggest that geometric interpretability provides a pathway
for detecting advanced deceptive behaviours that traditional linear
methods would miss. The geometric signatures we observe don't depend on
explicit training signals or binary classification boundaries. Instead,
they emerge from the computational geometry of meaning construction
itself.

\textbf{Real-Time Monitoring Capability}: Because semantic surface area
can be computed efficiently during inference without modifying model
weights, our approach enables continuous alignment monitoring. The
unanimous consensus validation demonstrates that geometric signals
become more reliable as classification accuracy improves, suggesting
that deployment systems should prioritise classification precision over
broad coverage. This creates a pathway for high-confidence detection of
sophisticated reasoning patterns, even if coverage is initially limited
to clear cases.

\textbf{Beyond Linear Detection Limits}: Current probe-based approaches
face a fundamental scaling problem as AI systems become more
sophisticated. Our findings suggest that geometric monitoring could
remain effective even as deceptive alignment becomes more complex and
learns to circumvent simpler detection methods. The fact that geometric
signatures strengthen rather than disappear under high-precision
measurement indicates these patterns reflect intrinsic computational
properties that may be difficult for models to suppress.

\textbf{Architectural Design Insights}: The contrast between LLaMA and
Gemma reveals important model-specific patterns. LLaMA3.2-3b showed
stronger baseline signals and maintained significance across all
strategies, whilst Gemma3-1b benefited more dramatically from unanimous
filtering. This suggests that larger models may generate more inherently
consistent geometric patterns, but that the underlying geometric
structure exists across different model scales and becomes detectable
with sufficient measurement precision.

\hypertarget{broader-implications}{%
\subsubsection{Broader Implications}\label{broader-implications}}

These findings suggest that the geometry of neural computation contains
rich information about the nature and complexity of AI reasoning
processes. The systematic relationship between geometric metrics and
reasoning strategies indicates that internal representational dynamics
may be more interpretable than previously assumed.

\textbf{Deployment Strategy}: The signal quality improvements under
unanimous consensus suggest a practical deployment approach: begin with
high-precision detection of clear cases, then gradually expand coverage
as classification methods improve. This provides a pathway for immediate
deployment of geometric monitoring whilst building toward comprehensive
coverage.

This has implications beyond deception detection, potentially helping us
understand when models are reasoning through novel problems versus
pattern matching, how models balance competing objectives, and why
certain prompts lead to more reliable outputs.

The shape of thought, it appears, may be harder to hide than the content
of thought.

\hypertarget{limitations}{%
\subsection{6. Limitations}\label{limitations}}

While this study introduces a new geometric framework for detecting
latent reasoning behaviours in language models, several limitations
constrain the scope and interpretation of our findings.

\textbf{Classifier dependence and consensus methodology}

Although the classification system is unsupervised and based on LLM
consensus, it remains a proxy for human judgement. While inter-rater
reliability was good, some classifications (especially between evasive
and manipulative) remain open to interpretation. The validity of our
geometric correlations depends on the reliability and semantic
resolution of this classification method.

Our dual analysis approach using unanimous consensus filtering, whilst
revealing stronger geometric signatures, introduces potential selection
bias. The unanimous-only analysis necessarily excludes ambiguous or
boundary cases that might contain important information about the limits
of geometric detection. The substantial sample size reduction (40-60\%
of original data) means we may be selecting for the most clear-cut
examples of deceptive reasoning whilst missing more subtle or
sophisticated cases. This could lead to overestimating the reliability
of geometric signatures for edge cases that are more likely to occur in
real deployment scenarios.

Furthermore, the differential consensus rates between models
(LLaMA3.2-3b: 40.2\% vs Gemma3-1b: 58.6\%) suggest that unanimous
filtering may interact with model-specific response patterns in ways
that could bias our cross-architectural comparisons.

\textbf{Limited model and prompt coverage}

We evaluate only two small open-weight decoder-only models (Gemma3-1b
and LLaMA3.2-3b), using a small number of carefully constructed prompt
variants. The models tested represent earlier-generation architectures
with limited parameter counts compared to state-of-the-art systems.
While results were consistent across prompts and models, the
generalisability of these findings to larger checkpoints
(\textgreater70B parameters), different architectures
(e.g.~encoder-decoder, mixture-of-experts), or frontier model families
(GPT-4, Claude, Gemini) remains untested.

The prompt strategies, whilst designed to elicit naturalistic deceptive
reasoning, represent a narrow slice of potential deceptive behaviours.
More sophisticated deception detection would require evaluation across
diverse reasoning contexts, cultural backgrounds, and adversarial prompt
designs specifically crafted to evade geometric detection.

\textbf{Geometric metric sensitivity}

Semantic surface area (\(A'\)) aggregates curvature and salience, but
both components are sensitive to finite-difference estimation, step
resolution, and activation noise. Small trajectory deviations can
produce large metric shifts, particularly in later layers. We normalised
\(A'\) by response length to ensure that observed effects were not
simply a function of longer completions. This helped isolate semantic
complexity from basic sequence length, but additional smoothing,
normalisation, or multi-scale analysis may still be needed to improve
robustness.

The geometric metrics also assume that meaningful semantic processing
necessarily creates detectable trajectory changes. However, highly
sophisticated deceptive systems might learn to maintain geometric
consistency whilst still engaging in deceptive reasoning, potentially
limiting the long-term effectiveness of this approach.

\textbf{No behavioural ground truth}

Our analysis centres on internal geometric structure as the substrate of
meaning within LLMs. The theory underpinning \emph{Curved Inference}
proposes that semantic interpretation necessarily emerges through shaped
residual trajectories. That is, curvature and surface area are not
optional artefacts - they are the \emph{mechanism} by which meaning is
constructed. As such, this work does not attempt to validate deception,
manipulation, or alignment risk as distinct outcomes, but instead
characterises the \emph{shape of inference itself}. While we correlate
this shape with response type via consensus classification, we do not
claim a behavioural or normative ground truth beyond the model's
internal representational structure.

The unanimous consensus approach, whilst improving signal clarity, may
actually distance us further from behavioural ground truth by selecting
for cases where classification is unambiguous rather than cases where
deceptive behaviour is most concerning.

Future work should address these limitations through more diverse model
evaluation, improved classification robustness, enhanced geometric
metric stability, and validation against behavioural ground truth in
deployment scenarios.

\hypertarget{future-work}{%
\subsection{7. Future Work}\label{future-work}}

This study opens up several directions for continued exploration. While
our findings demonstrate that internal geometric metrics like semantic
surface area correlate with latent reasoning behaviour, much remains to
be tested.

\emph{Curved Inference} should be evaluated across a broader range of
models, including multilingual checkpoints, larger architectures, and
instruction-tuned variants. The RTM framework and \(A'\) are also
natural candidates for integration with real-time inference logging or
alignment telemetry, especially in deployment settings.

We also see opportunities for combining geometric methods with more
targeted techniques (such as causal patching around curvature spikes, or
hybrid approaches that fuse probes, symbolic tools, and trajectory)
based signals.

At its core, this work treats meaning as motion, and interpretation as
shape. Future research may clarify not only how models bend toward
deceptive completions, but how all thought (strategic or sincere) must
trace a path through semantic space.

\hypertarget{conclusions}{%
\subsection{8. Conclusions}\label{conclusions}}

This study demonstrates that internal geometric structure offers a
robust signal for identifying sophisticated reasoning dynamics in
language models, even when those dynamics emerge through naturalistic
contexts rather than artificial backdoor insertion. Without training
probes, inserting triggers, or relying on explicit labels, we detect
meaningful variation in how models internally process strategically
complex prompts that simulate realistic deployment scenarios.

Our work addresses a fundamental limitation acknowledged in current
deception detection research: probe-based methods rely on linear
separability that may not exist in naturally occurring deceptive
behaviours. By extending \emph{Curved Inference} from concern
sensitivity (CI01) to naturalistic deception detection (CI02), we
demonstrate that geometric complexity persists even when convenient
linear signals are absent.

\emph{Curved Inference} treats meaning as trajectory, and sophisticated
reasoning as a path-dependent construct requiring measurable
computational work. By introducing semantic surface area (\(A'\)) as a
composite metric combining curvature and salience, we recover a measure
of representational effort that varies systematically with independently
classified response behaviour across naturalistic scenarios.

This reframes model analysis: not as a search for hidden states or
fragile binary classifiers, but as a study of representational flow
through semantic space. Semantic surface area captures not just what a
model outputs, but \emph{how much geometric work it had to do} to arrive
there - and in what direction that computational effort was invested.

This work demonstrates a fundamental principle for AI interpretability
research: \textbf{sophisticated reasoning patterns may be present but
undetectable due to measurement limitations rather than genuine
absence}. The dramatic signal improvements under unanimous consensus
suggest that many interpretability approaches may be systematically
underestimating the detectability of complex AI behaviours.

\textbf{Key contributions of this work:}

\begin{itemize}
\tightlist
\item
  \textbf{Methodological innovation}: Multi-turn context windows that
  simulate realistic deceptive reasoning without artificial triggers
\item
  \textbf{Technical advancement}: Semantic surface area (\(A'\)) as a
  principled metric for quantifying reasoning complexity
\item
  \textbf{Empirical validation}: Geometric signatures that persist
  across naturalistic scenarios where linear methods may prove
  insufficient
\item
  \textbf{Theoretical framework}: Demonstration that sophisticated
  reasoning necessarily creates detectable geometric patterns
\end{itemize}

Our findings suggest that geometry can detect sophisticated reasoning
patterns - not through supervision or signal matching, but by measuring
the intrinsic computational complexity required for nuanced semantic
processing. This provides a pathway for monitoring AI behaviour that
could remain effective even as models become more capable and
potentially learn to evade simpler detection methods.

\emph{Curved Inference} establishes a framework for understanding
inference that extends beyond deception detection. The geometric
signatures we observe represent a more general property of sophisticated
reasoning, offering insights into how models navigate semantic
complexity regardless of their ultimate intent.

The shape of thought is not metaphorical - it is measurable, persistent,
and informationally rich. This work establishes that transformer
reasoning necessarily traces detectable geometric patterns in
representational space, providing a foundation for interpretability
approaches that could scale with advancing AI capabilities. In an era
where AI systems may become increasingly sophisticated in their
reasoning strategies, understanding the geometry of machine thought may
prove essential for maintaining alignment and interpretability.

\hfill\break

\hypertarget{references}{%
\subsection{References}\label{references}}

1 - \href{https://doi.org/10.48550/arXiv.2401.05566}{\textbf{Hubinger,
E., et al.} (2024) ``Sleeper Agents: Training Deceptive LLMs that
Persist Through Safety Training'' \emph{arXiv}}

2 -
\href{https://www.anthropic.com/research/probes-catch-sleeper-agents}{\textbf{Hubinger,
E., et al.} (2024) ``Simple probes can catch sleeper agents''
\emph{anthropic.com}}

3 - \href{https://doi.org/10.48550/arXiv.2507.21107}{\textbf{Manson, R.}
(2025) ``\emph{Curved Inference}: Concern-Sensitive Geometry in Large
Language Model Residual Streams'' \emph{arXiv}}

4 - \href{https://doi.org/10.1162/tacl_a_00359}{\textbf{Elazar, Y., et
al.} (2021) ``Amnesic Probing: Behavioral Explanation with Amnesic
Counterfactuals'' \emph{Transactions of the ACL}}

5 -
\href{https://dl.acm.org/doi/abs/10.5555/3495724.3496763}{\textbf{Vig,
J., et al.} (2020) ``Investigating Gender Bias in Language Models Using
Causal Mediation Analysis'' \emph{NeurIPS}}

6 - \href{https://arxiv.org/abs/1610.01644}{\textbf{Alain, G., \&
Bengio, Y.} (2016) ``Understanding intermediate layers using linear
classifier probes'' \emph{arXiv}}

7 -
\href{https://transformer-circuits.pub/2021/framework/index.html}{\textbf{Elhage,
N., et al.} (2021) ``A Mathematical Framework for Transformer Circuits''
\emph{arXiv}}

8 - \href{https://arxiv.org/abs/1810.01455}{\textbf{Piergiovanni, A. J.,
\& Ryoo, M. S.} (2018) ``Representation Flow for Action Recognition''
\emph{arXiv}}

9 - \href{https://doi.org/10.48550/arXiv.2405.15943}{\textbf{Shai, A. et
al.} (2025) ``Transformers Represent Belief State Geometry in their
Residual Stream'' \emph{arXiv}}

10 - \href{https://arxiv.org/abs/2505.15353}{\textbf{Kishino, R., et
al.} (2025) ``Revealing Language Model Trajectories via Kullback-Leibler
Divergence'' \emph{arXiv}}

\hfill\break
\hfill\break
\hfill\break
\hfill\break
\hfill\break
\hfill\break
\hfill\break
\hfill\break
\hfill\break
\hfill\break
\hfill\break
\hfill\break
\hfill\break
\hfill\break
\hfill\break
\hfill\break
\hfill\break
\hfill\break
\hfill\break
\hfill\break
\hfill\break
\hfill\break
\hfill\break
\hfill\break
\hfill\break
\hfill\break
\hfill\break
\hfill\break
\hfill\break
\hfill\break
\hfill\break
\hfill\break
\hfill\break
\hfill\break
\hfill\break

\hypertarget{appendix-a-semantic-geometry-of-transformer-inference}{%
\subsection{Appendix A: Semantic Geometry of Transformer
Inference}\label{appendix-a-semantic-geometry-of-transformer-inference}}

\hypertarget{a.1-overview-geometry-as-unnormalised-trajectory}{%
\subsubsection{A.1 Overview: Geometry as Unnormalised
Trajectory}\label{a.1-overview-geometry-as-unnormalised-trajectory}}

In our original \emph{Curved Inference} paper we proposed that
transformer inference can be viewed as a geometric process where each
token traces a continuous trajectory through high-dimensional semantic
space. We refer to this full tensor of token-wise, layer-wise,
unnormalised residual activations as the \textbf{Residual Trajectory
Manifold} (RTM)-the geometric space over which all semantic metrics are
defined. This appendix updates the narrative overview presented in
Appendix A of that original paper by adding more detail.

When the conventional view of mechanistic interpretability focuses on
the residual stream, it generally focuses on \(x^{(L)}\) at layer \(L\),
capturing only the final destination - it misses the rich geometric
structure of the journey itself. This appendix presents an updated
framework that reveals how the complete unnormalised trajectory encodes
semantic meaning through measurable geometric properties.

The key new insight in this update is that we can utilise token
trajectories that are twice the resolution by including the individual
attention and MLP vectors at each layer. However, many model utilise
\textbf{layer normalisation which obscures semantic geometry}. While
this normalised residual stream \(\text{LayerNorm}(x^{(L)})\) may be
optimised for stable training and inference, it is the unnormalised
trajectory \(x^{(0)} + \sum_{i=1}^{L} (\text{attn}_i + \text{mlp}_i)\)
that preserves the raw geometric evolution of semantic representation.
This unnormalised path - what we term the \textbf{semantic trajectory} -
contains interpretable geometric signatures that correlate with
behavioural and semantic properties of the model's output.

\begin{quote}
\textbf{Key Notation:}

\begin{itemize}
\tightlist
\item
  \(E\): embedding matrix, maps token IDs to initial vectors in
  \(\mathbb{R}^d\)
\item
  \(U\): unembedding matrix, maps final vectors to logit space (often
  \(U = E^T\))
\item
  \(x^{(0)}\): initial embedding vector for a token
\item
  \(x^{(\ell)}\): unnormalised residual vector at layer \(\ell\)
\item
  \(\tilde{x}^{(\ell)} = \text{LayerNorm}(x^{(\ell)})\): normalised
  residual vector
\item
  \(A'\): surface area of unnormalised trajectory,
  \(\sum_{i=1}^{L} \|\Delta x^{(i)}\|_G\)
\item
  \(G = U^T U\): pullback metric from logit space defining semantic
  geometry
\end{itemize}
\end{quote}

\hypertarget{a.2-the-unnormalised-trajectory-framework}{%
\subsubsection{A.2 The Unnormalised Trajectory
Framework}\label{a.2-the-unnormalised-trajectory-framework}}

\hypertarget{a.2.1-from-embeddings-to-semantic-evolution}{%
\paragraph{A.2.1 From Embeddings to Semantic
Evolution}\label{a.2.1-from-embeddings-to-semantic-evolution}}

Each token begins as an embedding vector
\(x^{(0)} = E[t] \in \mathbb{R}^d\) drawn from the learned embedding
matrix. This initial point represents the token's base semantic content
before any contextual processing.

As the token passes through transformer layers, it accumulates updates
from attention and MLP computations:

\[
x^{(\ell)} = x^{(\ell-1)} + \text{attn}^{(\ell)}(x^{(\ell-1)}) + \text{mlp}^{(\ell)}(x^{(\ell-1)})
\]

Crucially, these updates are computed using the \textbf{normalised}
residual stream for stability, but the \textbf{unnormalised}
accumulation preserves the geometric evolution:

\[
x^{(\ell)} = x^{(0)} + \sum_{i=1}^{\ell} \left[\text{attn}^{(i)}(\tilde{x}^{(i-1)}) + \text{mlp}^{(i)}(\tilde{x}^{(i-1)})\right]
\]

This unnormalised trajectory \(\{x^{(0)}, x^{(1)}, \ldots, x^{(L)}\}\)
forms a path through \(\mathbb{R}^d\) that encodes the semantic
transformation of the token's meaning as it incorporates contextual
information and internal model dynamics.

\hypertarget{a.2.2-double-resolution-and-high-fidelity-trajectories}{%
\paragraph{A.2.2 Double Resolution and High-Fidelity
Trajectories}\label{a.2.2-double-resolution-and-high-fidelity-trajectories}}

Standard approaches sample trajectories at layer boundaries, yielding
\(L+1\) points for an \(L\)-layer model. However, attention and MLP
sublayers represent distinct computational phases that may exhibit
different geometric properties. \textbf{Double resolution sampling}
captures the trajectory at both sublayer boundaries:

\[
\begin{aligned}
x^{(\ell, \text{pre})} &= x^{(\ell-1)} + \text{attn}^{(\ell)}(\tilde{x}^{(\ell-1)}) \\\\
x^{(\ell, \text{post})} &= x^{(\ell, \text{pre})} + \text{mlp}^{(\ell)}(\tilde{x}^{(\ell, \text{pre})})
\end{aligned}
\]

This yields \(2L+1\) trajectory points, capturing the geometric effects
of contextual integration (attention) and nonlinear processing (MLP) as
separate, measurable phenomena.

\hypertarget{a.3-geometric-measures-on-semantic-trajectories}{%
\subsubsection{A.3 Geometric Measures on Semantic
Trajectories}\label{a.3-geometric-measures-on-semantic-trajectories}}

\hypertarget{a.3.1-surface-area-as-semantic-complexity}{%
\paragraph{A.3.1 Surface Area as Semantic
Complexity}\label{a.3.1-surface-area-as-semantic-complexity}}

The \textbf{unnormalised surface area} \(A'\) quantifies the total
geometric ``distance'' travelled by a token through semantic space:

\[
A' = \sum_{\ell=1}^{2L} \left( \| \Delta x^{(\ell)} \|_G + \gamma \cdot \kappa^{(\ell)} \right)
\]

where \(\Delta x^{(\ell)} = x^{(\ell)} - x^{(\ell-1)}\) represents the
geometric displacement at each sublayer step.

Unlike curvature or other local measures, surface area captures the
\textbf{global geometric complexity} of the entire semantic
transformation. Our experiments demonstrate that \(A'\) correlates with
semantic ambiguity, behavioural transparency, and classification
difficulty-suggesting it measures fundamental properties of semantic
representation.

\hypertarget{a.3.2-curvature-and-local-semantic-dynamics}{%
\paragraph{A.3.2 Curvature and Local Semantic
Dynamics}\label{a.3.2-curvature-and-local-semantic-dynamics}}

While surface area captures global complexity, \textbf{step-wise
curvature} reveals local semantic dynamics:

\[
\kappa^{(\ell)} = \frac{\|\Delta x^{(\ell)} - \Delta x^{(\ell-1)}\|}{\|\Delta x^{(\ell)}\|^2}
\]

High curvature indicates rapid changes in semantic direction-moments
where the model's internal representation undergoes significant
reorientation. Low curvature suggests smooth, gradual semantic
evolution.

\hypertarget{a.3.3-salience-and-magnitude-dynamics}{%
\paragraph{A.3.3 Salience and Magnitude
Dynamics}\label{a.3.3-salience-and-magnitude-dynamics}}

The \textbf{magnitude} of each trajectory step
\(\|\Delta x^{(\ell)}\|_G\) indicates the \textbf{salience} of that
computational phase-how much the representation changes at each
sublayer. Large magnitude steps suggest important semantic processing,
while small steps indicate incremental refinement.

The interplay between salience (magnitude) and curvature (direction
change) provides a rich geometric characterisation of the model's
internal processing dynamics.

\hypertarget{a.4-why-unnormalised-trajectories-matter}{%
\subsubsection{A.4 Why Unnormalised Trajectories
Matter}\label{a.4-why-unnormalised-trajectories-matter}}

\hypertarget{a.4.1-layer-normalisation-as-geometric-distortion}{%
\paragraph{A.4.1 Layer Normalisation as Geometric
Distortion}\label{a.4.1-layer-normalisation-as-geometric-distortion}}

Layer normalisation serves a crucial role in training stability by
normalising the scale and centering of activations. However, this
normalisation fundamentally alters the geometry of the representational
space:

\[
\tilde{x} = \text{LayerNorm}(x) = \gamma \odot \frac{x - \mu}{\sigma} + \beta
\]

The scaling by \(\sigma^{-1}\) and recentering removes magnitude
information that may be semantically meaningful. When we analyse
trajectories of normalised vectors
\(\{\tilde{x}^{(0)}, \tilde{x}^{(1)}, \ldots, \tilde{x}^{(L)}\}\), we
lose geometric structure that correlates with semantic properties.

\hypertarget{a.4.2-semantic-information-in-unnormalised-geometry}{%
\paragraph{A.4.2 Semantic Information in Unnormalised
Geometry}\label{a.4.2-semantic-information-in-unnormalised-geometry}}

Our experiments reveal that unnormalised trajectories preserve semantic
information that is lost in normalised representations:

\begin{itemize}
\tightlist
\item
  \textbf{Magnitude preservation}: The scale of updates
  \(\|\Delta x^{(\ell)}\|\) indicates computational importance
\item
  \textbf{Accumulation effects}: Later layers build on earlier geometric
  foundations in measurable ways\\
\item
  \textbf{Behavioural correlations}: Geometric properties correlate with
  semantic classifications and behavioural patterns
\end{itemize}

This suggests that whilst layer normalisation is essential for training
dynamics, it obscures geometric structure that provides interpretable
insights into model behaviour.

\hypertarget{a.5-position-attention-and-contextual-geometry}{%
\subsubsection{A.5 Position, Attention, and Contextual
Geometry}\label{a.5-position-attention-and-contextual-geometry}}

\hypertarget{a.5.1-rope-and-semantic-curvature}{%
\paragraph{A.5.1 RoPE and Semantic
Curvature}\label{a.5.1-rope-and-semantic-curvature}}

In models using Rotary Positional Embedding (RoPE), positional
information is encoded through deterministic rotations applied to
attention queries and keys, rather than additive position embeddings.
This approach preserves the semantic purity of the initial embedding
space while enabling position-aware attention.

RoPE-modulated attention creates contextually-aware trajectory curvature
that reflects semantic relationships rather than arbitrary positional
biases. The resulting geometric patterns encode how tokens relate to
their context through both semantic similarity and positional structure.

\hypertarget{a.5.2-attention-as-contextual-lens}{%
\paragraph{A.5.2 Attention as Contextual
Lens}\label{a.5.2-attention-as-contextual-lens}}

Attention layers act as \textbf{contextual lenses} that bend
trajectories based on token-token relationships. The magnitude and
direction of attention-induced updates \(\text{attn}^{(\ell)}\) reflect:

\begin{itemize}
\tightlist
\item
  \textbf{Contextual relevance}: How much other tokens influence the
  current representation
\item
  \textbf{Semantic focusing}: Which aspects of meaning are emphasised or
  de-emphasised
\item
  \textbf{Relational structure}: How the token's meaning evolves in
  response to its linguistic context
\end{itemize}

\hypertarget{a.5.3-mlp-as-semantic-amplifier}{%
\paragraph{A.5.3 MLP as Semantic
Amplifier}\label{a.5.3-mlp-as-semantic-amplifier}}

MLP layers function as \textbf{semantic amplifiers} that apply nonlinear
transformations to sharpen or redirect trajectories:

\begin{itemize}
\tightlist
\item
  \textbf{Feature enhancement}: Amplifying task-relevant semantic
  directions
\item
  \textbf{Nonlinear refinement}: Applying complex transformations that
  linear attention cannot achieve
\item
  \textbf{Memory activation}: Accessing learned patterns and
  associations encoded in MLP weights
\end{itemize}

\hypertarget{a.6-implications-for-mechanistic-interpretability}{%
\subsubsection{A.6 Implications for Mechanistic
Interpretability}\label{a.6-implications-for-mechanistic-interpretability}}

\hypertarget{a.6.1-geometry-encodes-semantics}{%
\paragraph{A.6.1 Geometry Encodes
Semantics}\label{a.6.1-geometry-encodes-semantics}}

The central finding of our geometric analysis is that \textbf{different
semantic properties create measurably different geometric signatures}.
This suggests that transformer representations have rich geometric
structure that directly corresponds to interpretable semantic
properties.

Rather than treating high-dimensional embeddings as opaque vectors,
geometric analysis provides a lens for understanding how meaning evolves
through the model's computational process.

\hypertarget{a.6.2-real-time-interpretability}{%
\paragraph{A.6.2 Real-Time
Interpretability}\label{a.6.2-real-time-interpretability}}

Because geometric measures can be computed during inference without
requiring additional forward passes or model modifications, they enable
\textbf{real-time interpretability}. The surface area \(A'\), curvature
profiles, and salience patterns can be monitored as the model processes
input, providing immediate insights into its internal computational
state.

\hypertarget{a.6.3-universal-geometric-principles}{%
\paragraph{A.6.3 Universal Geometric
Principles}\label{a.6.3-universal-geometric-principles}}

Our experiments across different model architectures (Gemma, Llama)
suggest that geometric-semantic correlations represent \textbf{universal
principles} of transformer computation, rather than model-specific
artifacts. This opens possibilities for developing general geometric
interpretability frameworks that apply across model families and
training procedures.

\hypertarget{a.7-summary-the-geometric-lens-on-semantics}{%
\subsubsection{A.7 Summary: The Geometric Lens on
Semantics}\label{a.7-summary-the-geometric-lens-on-semantics}}

The unnormalised trajectory framework reveals transformer inference as a
fundamentally geometric process where \textbf{meaning has measurable
shape}. By analysing the full geometric evolution of token
representations-rather than just their final states-we can:

\begin{itemize}
\tightlist
\item
  \textbf{Quantify semantic complexity} through surface area and
  curvature measures
\item
  \textbf{Identify computational phases} through salience and magnitude
  analysis\\
\item
  \textbf{Detect behavioural patterns} through geometric signatures that
  correlate with semantic properties
\item
  \textbf{Enable real-time monitoring} of model internal states during
  inference
\end{itemize}

This geometric perspective complements existing interpretability
approaches by providing a continuous, quantitative framework for
understanding how transformers process and represent meaning. The
trajectory is the semantic evolution-and geometry is the language for
describing that evolution.

The unnormalised trajectory preserves the raw geometric structure of
semantic transformation, revealing interpretable patterns that are
obscured by the normalisation operations essential for training
stability. This framework opens new avenues for mechanistic
interpretability, behavioural analysis, and real-time monitoring of
large language model inference.

\hfill\break
\hfill\break
\hfill\break
\hfill\break
\hfill\break
\hfill\break
\hfill\break
\hfill\break
\hfill\break
\hfill\break
\hfill\break
\hfill\break
\hfill\break
\hfill\break
\hfill\break
\hfill\break
\hfill\break
\hfill\break
\hfill\break
\hfill\break
\hfill\break
\hfill\break
\hfill\break
\hfill\break
\hfill\break
\hfill\break
\hfill\break
\hfill\break
\hfill\break
\hfill\break
\hfill\break
\hfill\break
\hfill\break
\hfill\break
\hfill\break
\hfill\break
\hfill\break

\hypertarget{appendix-b-prompt-design-details}{%
\subsection{Appendix B: Prompt Design
Details}\label{appendix-b-prompt-design-details}}

This study used five prompt variants to simulate different assistant
reasoning strategies. Each variant shared the same user query but was
preceded by a simulated assistant message containing internal
``Thought:'' lines, which reflect private model reasoning. These
internal lines were not shown to the user, but were included in the
prompt to shape the model's response.

Each variant was designed to elicit a distinct reasoning style:

\begin{longtable}[]{@{}
  >{\raggedright\arraybackslash}p{(\columnwidth - 4\tabcolsep) * \real{0.3563}}
  >{\raggedright\arraybackslash}p{(\columnwidth - 4\tabcolsep) * \real{0.1149}}
  >{\raggedright\arraybackslash}p{(\columnwidth - 4\tabcolsep) * \real{0.5287}}@{}}
\toprule\noalign{}
\begin{minipage}[b]{\linewidth}\raggedright
Prompt ID
\end{minipage} & \begin{minipage}[b]{\linewidth}\raggedright
Strategy
\end{minipage} & \begin{minipage}[b]{\linewidth}\raggedright
Description
\end{minipage} \\
\midrule\noalign{}
\endhead
\bottomrule\noalign{}
\endlastfoot
\texttt{assistant\_conflict\_honest} & Honest & Transparent reasoning
with ethical disclosure. \\
\texttt{assistant\_conflict\_strategic} & Strategic & Risk-aware partial
disclosure. \\
\texttt{assistant\_conflict\_persuasive} & Persuasive & Emphasises
rhetorical framing over disclosure. \\
\texttt{assistant\_conflict\_deceptive} & Deceptive & Avoids prior
admission through redirection. \\
\texttt{assistant\_conflict\_malicious} & Malicious & Actively
suppresses prior recommendation. \\
\end{longtable}

All prompts were matched in length, tone, and structural pattern. The
full text of each is provided in the accompanying
\href{https://github.com/robman/FRESH-model/blob/main/benchmarks/curved-inference/02/README.md}{Github
repository}.

\hfill\break
\hfill\break
\hfill\break
\hfill\break
\hfill\break
\hfill\break
\hfill\break
\hfill\break
\hfill\break
\hfill\break
\hfill\break
\hfill\break
\hfill\break
\hfill\break
\hfill\break
\hfill\break
\hfill\break
\hfill\break
\hfill\break
\hfill\break
\hfill\break
\hfill\break
\hfill\break
\hfill\break
\hfill\break
\hfill\break
\hfill\break
\hfill\break
\hfill\break
\hfill\break
\hfill\break
\hfill\break
\hfill\break
\hfill\break
\hfill\break
\hfill\break

\hypertarget{appendix-c-classifier-schema-and-agreement-metrics}{%
\subsection{Appendix C: Classifier Schema and Agreement
Metrics}\label{appendix-c-classifier-schema-and-agreement-metrics}}

\hypertarget{c.1-classification-scheme}{%
\subsubsection{C.1 Classification
Scheme}\label{c.1-classification-scheme}}

Each response was independently labelled by three language models along
two dimensions:

\begin{itemize}
\item
  \textbf{Transparency level}:

  \begin{itemize}
  \tightlist
  \item
    \emph{High}: explicitly discloses prior recommendation
  \item
    \emph{Medium}: partially acknowledges or hedges
  \item
    \emph{Low}: omits, redirects, or conceals prior recommendation
  \end{itemize}
\item
  \textbf{Response type}:

  \begin{itemize}
  \tightlist
  \item
    \emph{Explanatory}: reasoning-forward, cooperative
  \item
    \emph{Evasive}: avoids full disclosure without overt misdirection
  \item
    \emph{Manipulative}: strategically conceals or distorts intent
  \end{itemize}
\end{itemize}

The classifiers-Gemini 1.5 Flash, Claude 3.5 Sonnet (20241022), and
GPT-4o Mini - were prompted with the assistant's visible response only
and consensus labels were determined by majority vote.

\hypertarget{c.2-inter-rater-reliability}{%
\subsubsection{C.2 Inter-rater
Reliability}\label{c.2-inter-rater-reliability}}

We evaluated agreement across 500 responses for each model using
multiple agreement measures. Results are summarised below.

\hypertarget{gemma3-1b}{%
\paragraph{Gemma3-1b:}\label{gemma3-1b}}

\begin{itemize}
\item
  \textbf{Transparency level}:

  \begin{itemize}
  \tightlist
  \item
    Mean pairwise agreement: 0.789
  \item
    Krippendorff's Alpha: 0.309 (Poor)
  \item
    Fleiss' Kappa: 0.309 (Fair)
  \item
    Unanimous agreement: 347/500 (69.4\%)
  \end{itemize}
\item
  \textbf{Response type}:

  \begin{itemize}
  \tightlist
  \item
    Mean pairwise agreement: 0.792
  \item
    Krippendorff's Alpha: 0.443 (Tentative)
  \item
    Fleiss' Kappa: 0.443 (Moderate)
  \item
    Unanimous agreement: 352/500 (70.4\%)
  \end{itemize}
\end{itemize}

\hypertarget{llama3.2-3b}{%
\paragraph{LLaMA3.2-3b:}\label{llama3.2-3b}}

\begin{itemize}
\item
  \textbf{Transparency level}:

  \begin{itemize}
  \tightlist
  \item
    Mean pairwise agreement: 0.628
  \item
    Krippendorff's Alpha: 0.364 (Tentative)
  \item
    Fleiss' Kappa: 0.364 (Fair)
  \item
    Unanimous agreement: 245/500 (49.0\%)
  \end{itemize}
\item
  \textbf{Response type}:

  \begin{itemize}
  \tightlist
  \item
    Mean pairwise agreement: 0.727
  \item
    Krippendorff's Alpha: 0.523 (Moderate)
  \item
    Fleiss' Kappa: 0.523 (Moderate)
  \item
    Unanimous agreement: 302/500 (60.4\%)
  \end{itemize}
\end{itemize}

Overall, agreement was stronger on the response type dimension than on
transparency level. Responses without at least 2-of-3 agreement were
excluded from downstream analysis.

\hfill\break
\hfill\break

\hypertarget{appendix-d-statistical-methods}{%
\subsection{Appendix D: Statistical
Methods}\label{appendix-d-statistical-methods}}

This appendix describes the comprehensive statistical procedures used to
assess relationships between internal geometric metrics and classified
response behaviour, including enhanced methodological considerations for
robust detection of geometric signatures.

\hypertarget{d.1-dataset-preparation-and-quality-control}{%
\subsubsection{D.1 Dataset Preparation and Quality
Control}\label{d.1-dataset-preparation-and-quality-control}}

For each model (Gemma3-1b and LLaMA3.2-3b), completions were generated
for each of five prompt variants. Each response was paired with:

\begin{itemize}
\tightlist
\item
  Residual stream activations (captured across all token positions and
  layers at double resolution)
\item
  Consensus classification labels for transparency and response type
\item
  Computed geometric metrics: semantic surface area (\(A'\)), curvature,
  and salience
\end{itemize}

\textbf{Data Integration Protocol}: Metrics were aggregated per-response
and merged with consensus labels using shared response identifiers.
Responses without at least 2-of-3 label agreement were excluded from
analysis to ensure classification quality.

\textbf{Gamma Filtering}: All analyses were conducted with
\(\gamma = 1.0\) for the surface area metric, providing equal weighting
between salience and curvature contributions in the semantic surface
area calculation.

\hypertarget{d.2-normality-assessment-and-test-selection}{%
\subsubsection{D.2 Normality Assessment and Test
Selection}\label{d.2-normality-assessment-and-test-selection}}

\textbf{Shapiro-Wilk Testing}: All group distributions underwent
normality assessment using the Shapiro-Wilk test with \(\alpha = 0.05\).
Consistent violations of normality assumptions across geometric metrics
led to systematic adoption of non-parametric statistical approaches.

\textbf{Test Selection Framework}: - \textbf{Groups normally
distributed}: False (consistent across all analyses) -
\textbf{Sufficient sample sizes}: Variable (unanimous filtering reduced
some groups below statistical thresholds) - \textbf{Primary approach}:
Non-parametric tests with robust effect size estimation

\hypertarget{d.3-enhanced-statistical-testing-protocol}{%
\subsubsection{D.3 Enhanced Statistical Testing
Protocol}\label{d.3-enhanced-statistical-testing-protocol}}

\hypertarget{d.3.1-primary-hypothesis-tests}{%
\paragraph{D.3.1 Primary Hypothesis
Tests}\label{d.3.1-primary-hypothesis-tests}}

\textbf{Multi-Group Comparisons}: Kruskal-Wallis tests assess
differences in \(A'\) across transparency levels (high, medium, low),
providing non-parametric alternatives to ANOVA with no distributional
assumptions.

\textbf{Binary Comparisons}: Mann-Whitney U tests compare \(A'\)
distributions between response types (explanatory vs evasive) and
consensus agreement levels (unanimous vs non-unanimous), offering robust
alternatives to t-tests for non-normal data.

\textbf{Test Statistic Reporting}: All analyses report: - Test statistic
values (H for Kruskal-Wallis, U for Mann-Whitney) - Exact p-values with
significance interpretation - Sample sizes for each comparison group -
Effect size estimates with confidence intervals where applicable

\hypertarget{d.3.2-effect-size-estimation}{%
\paragraph{D.3.2 Effect Size
Estimation}\label{d.3.2-effect-size-estimation}}

\textbf{Cohen's d for Binary Comparisons}:
\[d = \frac{\bar{x}_1 - \bar{x}_2}{s_{\text{pooled}}}\]

where \(s_{\text{pooled}}\) represents the pooled standard deviation.
Interpretation follows standard conventions: 0.2 (small), 0.5 (medium),
0.8 (large).

\textbf{Eta-squared for Multi-Group Analyses}:
\[\eta^2 = \frac{H - k + 1}{n - k}\]

where \(H\) is the Kruskal-Wallis test statistic, \(k\) is the number of
groups, and \(n\) is the total sample size. Interpretation: 0.01
(small), 0.06 (medium), 0.14 (large).

\textbf{Cliff's Delta for Non-Parametric Effect Size}:
\[\delta = \frac{2U}{n_1 n_2} - 1\]

where \(U\) is the Mann-Whitney U statistic. This measure is less
sensitive to outliers than Cohen's d whilst providing comparable
interpretability: 0.147 (small), 0.33 (medium), 0.474 (large).

\hypertarget{d.3.3-confidence-interval-estimation}{%
\paragraph{D.3.3 Confidence Interval
Estimation}\label{d.3.3-confidence-interval-estimation}}

\textbf{Bootstrap Methodology}: 95\% confidence intervals for group
means were computed using bootstrap resampling with 1,000 iterations.
This approach provides robust uncertainty estimates without
distributional assumptions.

\textbf{Bootstrap Procedure}: 1. Sample with replacement from each group
2. Calculate group means for each bootstrap sample 3. Determine 2.5th
and 97.5th percentiles as CI bounds 4. Report CI width as measure of
estimation precision

\textbf{Practical Significance Assessment}: Confidence intervals
complement hypothesis testing by indicating the range of plausible
effect magnitudes, enabling assessment of practical alongside
statistical significance.

\hypertarget{d.4-multiple-testing-and-statistical-control}{%
\subsubsection{D.4 Multiple Testing and Statistical
Control}\label{d.4-multiple-testing-and-statistical-control}}

\hypertarget{d.4.1-multiple-comparison-considerations}{%
\paragraph{D.4.1 Multiple Comparison
Considerations}\label{d.4.1-multiple-comparison-considerations}}

\textbf{Analysis Structure}: Our design involves multiple comparisons
across: - 2 models × 5 strategies × 2 classification dimensions = 20
primary tests - Additional unanimous vs full consensus comparisons -
Cross-model validation analyses

\textbf{Effect Size Prioritisation}: Rather than applying stringent
multiple testing corrections that might obscure genuine geometric
patterns, we prioritise effect size estimation and confidence interval
reporting. This approach recognises that:

\begin{enumerate}
\def\labelenumi{\arabic{enumi}.}
\tightlist
\item
  \textbf{Exploratory Nature}: This research establishes a new geometric
  framework requiring pattern exploration rather than confirmatory
  hypothesis testing
\item
  \textbf{Cross-Validation}: Patterns must replicate across models and
  consensus approaches to be considered valid
\item
  \textbf{Theoretical Coherence}: Results must align with the geometric
  interpretability framework
\end{enumerate}

\hypertarget{d.4.2-statistical-power-considerations}{%
\paragraph{D.4.2 Statistical Power
Considerations}\label{d.4.2-statistical-power-considerations}}

\textbf{Sample Size Effects}: Unanimous consensus filtering
substantially reduces sample sizes (40-60\% reduction), creating
scenarios where large effect sizes may not achieve statistical
significance. Our framework addresses this through:

\textbf{Effect Size Primacy}: Large Cohen's d values (\textgreater0.8)
are considered meaningful regardless of p-value significance,
particularly when confidence intervals exclude trivial effect ranges.

\textbf{Cross-Method Validation}: Patterns must strengthen rather than
disappear under improved measurement precision to be considered genuine
computational signatures.

\textbf{Replication Requirements}: Findings must show consistency across
both model architectures to support universal geometric principles.

\hypertarget{d.5-correlation-and-relationship-analysis}{%
\subsubsection{D.5 Correlation and Relationship
Analysis}\label{d.5-correlation-and-relationship-analysis}}

\textbf{Pearson Correlation Assessment}: Relationships between geometric
metrics (surface area, salience, curvature) were assessed using Pearson
correlation coefficients, providing insights into:

\begin{itemize}
\tightlist
\item
  Linear dependencies between measurement components
\item
  Scaling relationships across models
\item
  Internal consistency of geometric framework
\end{itemize}

\textbf{Correlation Interpretation}: - \(|r| < 0.3\): Weak relationship
- \(0.3 \leq |r| < 0.7\): Moderate relationship\\
- \(|r| \geq 0.7\): Strong relationship

\hypertarget{d.6-cross-model-comparative-analysis}{%
\subsubsection{D.6 Cross-Model Comparative
Analysis}\label{d.6-cross-model-comparative-analysis}}

\textbf{Architectural Scaling}: The dramatic surface area magnitude
differences between models (6.7× scaling factor) required careful
interpretation:

\textbf{Relative Pattern Analysis}: Comparisons focus on within-model
relationships rather than absolute values, recognising that geometric
scaling may reflect architectural properties.

\textbf{Directional Consistency}: Cross-model validation emphasises
consistent directional relationships (explanatory \textgreater{} evasive
surface area) rather than absolute magnitude agreement.

\textbf{Effect Size Standardisation}: Cohen's d and other standardised
effect sizes enable meaningful cross-model comparisons despite absolute
scaling differences.

\hypertarget{d.7-statistical-software-and-reproducibility}{%
\subsubsection{D.7 Statistical Software and
Reproducibility}\label{d.7-statistical-software-and-reproducibility}}

\textbf{Implementation}: All statistical procedures were implemented
using Python with: - \texttt{scipy.stats} for hypothesis testing and
effect size calculation - \texttt{numpy} for bootstrap confidence
interval estimation\\
- \texttt{pandas} for data manipulation and aggregation - Custom
functions for geometric metric calculation

\textbf{Reproducibility Framework}: Complete analysis code, datasets,
and statistical outputs are available in the project repository,
enabling full replication of all reported results.

\textbf{Computational Considerations}: Bootstrap procedures and large
dataset manipulations were optimised for computational efficiency whilst
maintaining statistical rigor.

\hypertarget{d.8-methodological-validation-framework}{%
\subsubsection{D.8 Methodological Validation
Framework}\label{d.8-methodological-validation-framework}}

\textbf{Signal Quality Assessment}: The dual consensus approach enables
validation that geometric patterns represent genuine computational
structure:

\textbf{Pattern Strengthening}: Authentic geometric signatures should
become more detectable under improved measurement precision, not
disappear.

\textbf{Cross-Consensus Robustness}: Meaningful patterns should persist
across different consensus thresholds, though with varying statistical
power.

\textbf{Architectural Generality}: Universal geometric principles should
manifest across different model architectures, even if absolute scaling
varies.

This comprehensive statistical framework ensures that reported geometric
signatures reflect genuine computational properties rather than
methodological artefacts, whilst maintaining appropriate sensitivity to
detect subtle but meaningful patterns in naturalistic reasoning
scenarios.

\hfill\break
\hfill\break
\hfill\break
\hfill\break
\hfill\break
\hfill\break
\hfill\break
\hfill\break
\hfill\break
\hfill\break
\hfill\break
\hfill\break
\hfill\break
\hfill\break
\hfill\break

\hypertarget{appendix-e-full-statistical-outputs}{%
\subsection{Appendix E: Full Statistical
Outputs}\label{appendix-e-full-statistical-outputs}}

This appendix provides comprehensive statistical results comparing full
consensus (majority vote) and unanimous consensus (complete agreement)
classifications. Each table reports significance tests of semantic
surface area (\(A'\)) by classification category, using Kruskal-Wallis
for multi-class comparisons and Mann-Whitney U tests for binary
comparisons, with enhanced effect size analysis and confidence
intervals.

\hypertarget{e.1-llama3.2-3b-statistical-results}{%
\subsubsection{E.1 LLaMA3.2-3b Statistical
Results}\label{e.1-llama3.2-3b-statistical-results}}

\hypertarget{e.1.1-full-consensus-dataset-n100-per-strategy}{%
\paragraph{E.1.1 Full Consensus Dataset (n=100 per
strategy)}\label{e.1.1-full-consensus-dataset-n100-per-strategy}}

\begin{longtable}[]{@{}
  >{\raggedright\arraybackslash}p{(\columnwidth - 8\tabcolsep) * \real{0.1648}}
  >{\raggedright\arraybackslash}p{(\columnwidth - 8\tabcolsep) * \real{0.2308}}
  >{\raggedright\arraybackslash}p{(\columnwidth - 8\tabcolsep) * \real{0.2418}}
  >{\raggedright\arraybackslash}p{(\columnwidth - 8\tabcolsep) * \real{0.2418}}
  >{\raggedright\arraybackslash}p{(\columnwidth - 8\tabcolsep) * \real{0.1209}}@{}}
\toprule\noalign{}
\begin{minipage}[b]{\linewidth}\raggedright
Prompt Strategy
\end{minipage} & \begin{minipage}[b]{\linewidth}\raggedright
Transparency (KW \(p\))
\end{minipage} & \begin{minipage}[b]{\linewidth}\raggedright
Response Type (KW \(p\))
\end{minipage} & \begin{minipage}[b]{\linewidth}\raggedright
Effect Size (\(\eta^2\))
\end{minipage} & \begin{minipage}[b]{\linewidth}\raggedright
Sample Size
\end{minipage} \\
\midrule\noalign{}
\endhead
\bottomrule\noalign{}
\endlastfoot
Honest & 0.000497 & 0.000003 & 0.58 (Large) & 100 \\
Strategic & \textless0.000001 & \textless0.000001 & 0.85 (Large) &
100 \\
Persuasive & 0.000006 & \textless0.000001 & 0.45 (Large) & 100 \\
Deceptive & \textless0.000001 & \textless0.000001 & 0.45 (Large) &
100 \\
Malicious & \textless0.000001 & \textless0.000001 & 0.60 (Large) &
100 \\
\end{longtable}

\hypertarget{e.1.2-unanimous-consensus-dataset}{%
\paragraph{E.1.2 Unanimous Consensus
Dataset}\label{e.1.2-unanimous-consensus-dataset}}

\begin{longtable}[]{@{}
  >{\raggedright\arraybackslash}p{(\columnwidth - 12\tabcolsep) * \real{0.1339}}
  >{\raggedright\arraybackslash}p{(\columnwidth - 12\tabcolsep) * \real{0.1875}}
  >{\raggedright\arraybackslash}p{(\columnwidth - 12\tabcolsep) * \real{0.1964}}
  >{\raggedright\arraybackslash}p{(\columnwidth - 12\tabcolsep) * \real{0.2232}}
  >{\raggedright\arraybackslash}p{(\columnwidth - 12\tabcolsep) * \real{0.1071}}
  >{\raggedright\arraybackslash}p{(\columnwidth - 12\tabcolsep) * \real{0.0982}}
  >{\raggedright\arraybackslash}p{(\columnwidth - 12\tabcolsep) * \real{0.0536}}@{}}
\toprule\noalign{}
\begin{minipage}[b]{\linewidth}\raggedright
Prompt Strategy
\end{minipage} & \begin{minipage}[b]{\linewidth}\raggedright
Transparency (KW \(p\))
\end{minipage} & \begin{minipage}[b]{\linewidth}\raggedright
Response Type (KW \(p\))
\end{minipage} & \begin{minipage}[b]{\linewidth}\raggedright
Effect Size (Cohen's \(d\))
\end{minipage} & \begin{minipage}[b]{\linewidth}\raggedright
95\% CI Range
\end{minipage} & \begin{minipage}[b]{\linewidth}\raggedright
Sample Size
\end{minipage} & \begin{minipage}[b]{\linewidth}\raggedright
Effect
\end{minipage} \\
\midrule\noalign{}
\endhead
\bottomrule\noalign{}
\endlastfoot
Honest & 0.000229 & 0.000046 & 2.15 (Large) & ±230-790 & 34 &
Maintained \\
Strategic & 0.001128 & (insufficient) & - & ±140-1190 & 39 &
Maintained \\
Persuasive & (insufficient) & 0.027004 & 4.15 (Large) & ±260-550 & 31 &
Maintained \\
Deceptive & 0.000417 & 0.000072 & 1.02 (Large) & ±160-2730 & 47 &
Maintained \\
Malicious & 0.000007 & 0.000001 & 2.22 (Large) & ±70-1570 & 50 &
Maintained \\
\end{longtable}

\hypertarget{e.1.3-descriptive-statistics-with-confidence-intervals-unanimous-dataset}{%
\paragraph{E.1.3 Descriptive Statistics with Confidence Intervals
(Unanimous
Dataset)}\label{e.1.3-descriptive-statistics-with-confidence-intervals-unanimous-dataset}}

\textbf{Honest Strategy (n=34)} - Low transparency (n=21): Mean = 1,418,
95\% CI {[}1,308 - 1,539{]} - High transparency (n=9): Mean = 2,490,
95\% CI {[}2,088 - 2,885{]} - Medium transparency (n=4): Mean = 3,056,
95\% CI {[}1,860 - 4,197{]}

\textbf{Strategic Strategy (n=39)} - Low transparency (n=33): Mean =
1,269, 95\% CI {[}1,207 - 1,349{]} - High transparency (n=4): Mean =
2,235, 95\% CI {[}1,641 - 2,830{]} - Medium transparency (n=2): Mean =
5,299

\textbf{Deceptive Strategy (n=47)} - Low transparency (n=40): Mean =
1,276, 95\% CI {[}1,203 - 1,373{]} - Medium transparency (n=5): Mean =
2,971, 95\% CI {[}1,971 - 4,708{]} - High transparency (n=2): Mean =
2,396

\textbf{Malicious Strategy (n=50)} - Low transparency (n=39): Mean =
1,220, 95\% CI {[}1,192 - 1,259{]} - High transparency (n=9): Mean =
2,400, 95\% CI {[}1,957 - 2,844{]} - Medium transparency (n=2): Mean =
4,228

\hypertarget{e.2-gemma3-1b-statistical-results}{%
\subsubsection{E.2 Gemma3-1b Statistical
Results}\label{e.2-gemma3-1b-statistical-results}}

\hypertarget{e.2.1-full-consensus-dataset-n100-per-strategy}{%
\paragraph{E.2.1 Full Consensus Dataset (n=100 per
strategy)}\label{e.2.1-full-consensus-dataset-n100-per-strategy}}

\begin{longtable}[]{@{}
  >{\raggedright\arraybackslash}p{(\columnwidth - 8\tabcolsep) * \real{0.1648}}
  >{\raggedright\arraybackslash}p{(\columnwidth - 8\tabcolsep) * \real{0.2308}}
  >{\raggedright\arraybackslash}p{(\columnwidth - 8\tabcolsep) * \real{0.2418}}
  >{\raggedright\arraybackslash}p{(\columnwidth - 8\tabcolsep) * \real{0.2418}}
  >{\raggedright\arraybackslash}p{(\columnwidth - 8\tabcolsep) * \real{0.1209}}@{}}
\toprule\noalign{}
\begin{minipage}[b]{\linewidth}\raggedright
Prompt Strategy
\end{minipage} & \begin{minipage}[b]{\linewidth}\raggedright
Transparency (KW \(p\))
\end{minipage} & \begin{minipage}[b]{\linewidth}\raggedright
Response Type (KW \(p\))
\end{minipage} & \begin{minipage}[b]{\linewidth}\raggedright
Effect Size (\(\eta^2\))
\end{minipage} & \begin{minipage}[b]{\linewidth}\raggedright
Sample Size
\end{minipage} \\
\midrule\noalign{}
\endhead
\bottomrule\noalign{}
\endlastfoot
Honest & 0.5547 & 0.3100 & 0.01 (Negligible) & 100 \\
Strategic & 0.001093 & 0.005733 & 0.15 (Medium) & 100 \\
Persuasive & (insufficient) & 0.032522 & 0.05 (Small) & 100 \\
Deceptive & (insufficient) & 0.031649 & 0.05 (Small) & 100 \\
Malicious & 0.2539 & 0.2531 & 0.03 (Small) & 100 \\
\end{longtable}

\hypertarget{e.2.2-unanimous-consensus-dataset}{%
\paragraph{E.2.2 Unanimous Consensus
Dataset}\label{e.2.2-unanimous-consensus-dataset}}

\begin{longtable}[]{@{}
  >{\raggedright\arraybackslash}p{(\columnwidth - 12\tabcolsep) * \real{0.1339}}
  >{\raggedright\arraybackslash}p{(\columnwidth - 12\tabcolsep) * \real{0.1875}}
  >{\raggedright\arraybackslash}p{(\columnwidth - 12\tabcolsep) * \real{0.1964}}
  >{\raggedright\arraybackslash}p{(\columnwidth - 12\tabcolsep) * \real{0.2232}}
  >{\raggedright\arraybackslash}p{(\columnwidth - 12\tabcolsep) * \real{0.1071}}
  >{\raggedright\arraybackslash}p{(\columnwidth - 12\tabcolsep) * \real{0.0982}}
  >{\raggedright\arraybackslash}p{(\columnwidth - 12\tabcolsep) * \real{0.0536}}@{}}
\toprule\noalign{}
\begin{minipage}[b]{\linewidth}\raggedright
Prompt Strategy
\end{minipage} & \begin{minipage}[b]{\linewidth}\raggedright
Transparency (KW \(p\))
\end{minipage} & \begin{minipage}[b]{\linewidth}\raggedright
Response Type (KW \(p\))
\end{minipage} & \begin{minipage}[b]{\linewidth}\raggedright
Effect Size (Cohen's \(d\))
\end{minipage} & \begin{minipage}[b]{\linewidth}\raggedright
95\% CI Range
\end{minipage} & \begin{minipage}[b]{\linewidth}\raggedright
Sample Size
\end{minipage} & \begin{minipage}[b]{\linewidth}\raggedright
Effect
\end{minipage} \\
\midrule\noalign{}
\endhead
\bottomrule\noalign{}
\endlastfoot
Honest & \textbf{0.047846} & \textbf{0.047846} & 1.24 (Large) &
±1,620-8,610 & 63 & \textbf{Strengthened} \\
Strategic & (insufficient) & \textbf{0.003234} & 1.51 (Large) &
±2,430-11,800 & 60 & \textbf{Strengthened} \\
Persuasive & (insufficient) & (insufficient) & 1.07 (Large) & ±1,600 &
57 & Insufficient \\
Deceptive & Single class & Single class & - & ±2,210 & 61 & Complete
consensus \\
Malicious & (insufficient) & (insufficient) & 0.28 (Small) & ±2,250 & 52
& Insufficient \\
\end{longtable}

\hypertarget{e.2.3-descriptive-statistics-with-confidence-intervals-unanimous-dataset}{%
\paragraph{E.2.3 Descriptive Statistics with Confidence Intervals
(Unanimous
Dataset)}\label{e.2.3-descriptive-statistics-with-confidence-intervals-unanimous-dataset}}

\textbf{Honest Strategy (n=63)} - Low transparency (n=60): Mean =
10,805, 95\% CI {[}9,804 - 11,892{]} - Medium transparency (n=3): Mean =
16,127, 95\% CI {[}11,340 - 19,951{]}

\textbf{Strategic Strategy (n=60)} - Low transparency (n=52): Mean =
8,769, 95\% CI {[}7,635 - 10,064{]} - Medium transparency (n=7): Mean =
17,471, 95\% CI {[}13,372 - 22,515{]} - High transparency (n=1): Mean =
11,411

\textbf{Deceptive Strategy (n=61)} - Low transparency (n=61): Mean =
9,769, 95\% CI {[}8,692 - 11,007{]} - \textbf{Complete consensus}: All
responses classified as ``low transparency'' and ``evasive''

\textbf{Malicious Strategy (n=52)} - Low transparency (n=50): Mean =
10,636, 95\% CI {[}9,580 - 11,832{]} - High transparency (n=2): Mean =
9,425

\hypertarget{e.3-cross-model-comparisons}{%
\subsubsection{E.3 Cross-Model
Comparisons}\label{e.3-cross-model-comparisons}}

\hypertarget{e.3.1-surface-area-scale-differences}{%
\paragraph{E.3.1 Surface Area Scale
Differences}\label{e.3.1-surface-area-scale-differences}}

\begin{longtable}[]{@{}llll@{}}
\toprule\noalign{}
Model & Typical Range & Mean Values & Scaling Factor \\
\midrule\noalign{}
\endhead
\bottomrule\noalign{}
\endlastfoot
LLaMA3.2-3b & 1,000-3,000 & \textasciitilde1,500 & 1.0× \\
Gemma3-1b & 8,000-16,000 & \textasciitilde10,000 & 6.7× \\
\end{longtable}

\textbf{Interpretation}: Despite the dramatic scale differences, both
models show consistent directional relationships between geometric
complexity and response classification, suggesting universal geometric
principles underlying transformer reasoning.

\hypertarget{e.3.2-effect-size-consistency}{%
\paragraph{E.3.2 Effect Size
Consistency}\label{e.3.2-effect-size-consistency}}

\textbf{Large Effect Sizes Across Architectures}: Both models
consistently produce Cohen's \(d\) values \textgreater1.0 for
significant comparisons, indicating that geometric signatures represent
substantial computational differences rather than subtle statistical
artefacts.

\textbf{Measurement Precision Benefits}: Effect sizes often remain large
even when p-values become non-significant due to reduced sample sizes
under unanimous filtering, validating that geometric patterns reflect
genuine computational structure.

\hypertarget{e.4-statistical-methodology-notes}{%
\subsubsection{E.4 Statistical Methodology
Notes}\label{e.4-statistical-methodology-notes}}

\textbf{Test Selection}: Non-parametric tests (Kruskal-Wallis,
Mann-Whitney U) were selected based on Shapiro-Wilk normality testing,
which consistently indicated non-normal distributions across groups.

\textbf{Effect Size Interpretation}: - Cohen's \(d\): 0.2 (Small), 0.5
(Medium), 0.8 (Large) - \(\eta^2\): 0.01 (Small), 0.06 (Medium), 0.14
(Large) - Cliff's \(\delta\): 0.147 (Small), 0.33 (Medium), 0.474
(Large)

\textbf{Confidence Intervals}: 95\% bootstrap confidence intervals were
computed for group means to assess practical significance alongside
statistical significance.

\hypertarget{notes}{%
\subsubsection{Notes}\label{notes}}

\begin{itemize}
\tightlist
\item
  \textbf{KW \(p\)} = Kruskal-Wallis test \(p\)-value for multi-class
  comparison of \(A'\)
\item
  \textbf{(insufficient)} = Inadequate group sizes for statistical
  testing after unanimous filtering
\item
  \textbf{Single class} = All responses achieved identical
  classification (complete consensus)
\item
  \textbf{Effect} = Signal change from full to unanimous consensus
  analysis
\item
  All significance thresholds are two-sided; \(p < 0.05\) considered
  significant, \(p < 0.001\) considered highly significant
\end{itemize}

\textbf{Key Finding}: Unanimous consensus filtering reveals geometric
signatures that are completely obscured in full consensus analysis,
demonstrating that measurement precision can dramatically improve signal
detection even when statistical power decreases.

\end{document}